\RequirePackage{fix-cm}
\documentclass[smallextended]{svjour3}       
\usepackage{preamble}
\usepackage{natbib}

\makeatletter
\def\Cline#1#2{\@Cline#1#2\@nil}
\def\@Cline#1-#2#3\@nil{%
  \omit
  \@multicnt#1%
  \advance\@multispan\m@ne
  \ifnum\@multicnt=\@ne\@firstofone{&\omit}\fi
  \@multicnt#2%
  \advance\@multicnt-#1%
  \advance\@multispan\@ne
  \leaders\hrule\@height#3\hfill
  \cr}
\makeatother

\begin{document}

\title{Modelling Serendipity in a Computational  Context\thanks{This research was supported by the Engineering and Physical Sciences Research Council through grants EP/L00206X, EP/J004049, EP/L015846/1, and EP/K040251/2, with additional workshop support through EP/N014758/1; by the Future and Emerging Technologies (FET) programme within the Seventh Framework Programme for Research of the European Commission, under FET-Open Grant numbers: 611553 (COINVENT) and 611560 (WHIM); and by workshop and symposium support from the Society for the study of Artificial Intelligence and Simulation of Behaviour(AISB).}}



\author{Joseph Corneli \and Anna Jordanous \and Christian Guckelsberger \and Alison Pease \and Simon Colton}


\institute{J. Corneli \at
  Hyperreal Enterprises, Ltd.
  \email{joseph.corneli@hyperreal.enterprises}\\
  ORCID: 0000-0003-1330-4698
  \and
  A. Jordanous \at
  School of Computing, University of Kent, Chatham Maritime ME4 4AG, UK.
  \email{a.k.jordanous@kent.ac.uk} (Corresponding author)\\
  ORCID: 0000-0003-2076-8642
  \and
  C. Guckelsburger;  S. Colton  \at
  School of Electronic Engineering and Computer Science, Queen Mary University of London, Mile End Road, London E1 4NS, UK.
  \email{christian.guckelsberger@qmul.ac.uk; s.colton@qmul.ac.uk}\\
  ORCID: 0000-0003-1977-1887 (CG); 0000-0003-3377-1680 (SC)
  \and
  A. Pease  \at
  School of Science \& Engineering, University of Dundee, Dundee DD1 4HN, UK.
  \email{a.pease@dundee.ac.uk}\\
  ORCID: 0000-0003-1856-9599
  }

\date{Received: date / Accepted: date}

\maketitle

\begin{abstract}  
The term serendipity describes a creative process that
develops, in context, with the active participation of a creative
agent, but not entirely within that agent's control.
While a system cannot be made to perform serendipitously on demand,
we argue that its \emph{serendipity potential} can be
increased by means of a suitable system architecture and other design choices.
We distil a unified description of serendipitous occurrences from
historical theorisations of serendipity and creativity.  This takes
the form of a framework with six phases: \emph{perception},
\emph{attention}, \emph{interest}, \emph{explanation}, \emph{bridge},
and \emph{valuation}.
We then use this framework to organise a survey 
of literature in cognitive science, philosophy, and computing,
which yields practical definitions
of the six phases, along with heuristics for implementation.
We use the resulting model
to evaluate the serendipity potential of four existing systems
developed by others, and two systems previously developed by two of the authors.
Most existing
research that considers serendipity in a computing context deals with
serendipity as a service; here we relate
theories of serendipity to the development
of autonomous systems and computational creativity practice.
We argue that serendipity is not teleologically blind, and outline representative directions for future applications of our model.
We conclude that it is feasible to equip computational systems with
the potential for serendipity, and that this could be beneficial in
varied computational creativity/AI applications, particularly those
designed to operate responsively in real-world contexts.
\end{abstract}

\ifdraft{\tableofcontents
\clearpage
}{}

\section{Introduction} \label{sec:introduction}

Serendipity has played a role in many human discoveries: often-cited examples
range from vulcanized rubber, the
Velcro\textsuperscript{\texttrademark} strip, and 3M's ubiquitous
Post-it\textsuperscript{\textregistered} Notes, through to penicillin,
LSD, and Viagra\textsuperscript{\textregistered}.  
An improved understanding of
serendipity could help bring about (computationally) creative
breakthroughs in these areas.

Given its crucial role in human discovery and invention, it is not surprising that the concept of serendipity has been adopted for users' benefit by many research areas such as computational creativity \citep{pease2013discussion}, information retrieval \citep{Toms2000, Andre:2009:XSP:1518701.1519009}, recommender systems \citep{kotkov2016survey,Zhang2011}, creativity support tools \cite{maxwell2012designing} and planning \citep{muscettola1997board, chakraborti2015planning}. 
Crucially, all of these examples use the concept of serendipity to denote and design systems which stimulate the experience of serendipity in their users - what we term: \emph{serendipity as a service}. Here, we propose to switch perspectives from ``serendipity as a service'' to ``\emph{serendipity in the system},'' where artificial systems can catalyse, evaluate and leverage serendipitous occurrences themselves.

This perspective shift requires a more nuanced understanding of serendipity: for example, consider a reversal of roles in which a person contributes to a system's experience of serendipity, in some suitable sense.  Here our central goal is to theorise, and indicate in broad terms how to engineer, systems which do not depend on such support by people, but which have the capacity to detect, evaluate and use serendipitous events without  user intervention.  Why might such features be useful?  \citet{delamaza1994generate} raised the point: ``How disastrous it would be if a discovery system's greatest discovery was `not noticed' because a human did not have the ability to recognise it!''



Contrary to de la Maza's hopes, van Andel has suggested that an artificial system could never be independent of a person in leveraging serendipity.
\begin{quote}
\emph{``Like all intuitive operating, pure serendipity is not amenable to generation by a computer. The very moment I can plan or programme `serendipity' it cannot be called serendipity anymore. All I can programme is, that, if the unforeseen happens, the system alerts the user and incites him to observe and act by himself by trying to make a correct abduction of the surprising fact or relation.''}  \cite[p.~646]{van1994anatomy}
\end{quote}
We fully agree that an artificial system cannot be guaranteed to engage in serendipitous findings, just as a person cannot deliberately force serendipity to happen ``on demand.''  However, we argue that serendipity can happen independently of human intervention within an artificial system, and that the ``\emph{serendipity potential}'' of such a system can be increased by means of a suitable system architecture. 
In a comparable human context, Louis Pasteur (known for serendipitous discoveries in chemistry and biology \citep{roberts,gaughan2010accidental}) famously remarked: ``Dans les champs de l'observation le hasard ne favorise que les esprits préparés'' (``In the fields of observation chance favours only prepared minds'') \cite[][p. 131]{pasteur-chance}.\footnote{Van Andel pointed out (p.c.) that Pasteur's manuscript actually says ``Dans les champs de l’observation,
le hasard ne favorise que \emph{des} esprits pr\'epar\'es'' \citep{bourcier2011serendipite}---``In the fields of observation chance favours only \emph{some} prepared minds.''}  ``Preparedness'' encompasses various ways in which the serendipity potential of a system can be enhanced.  

The framework that we advance was inspired by earlier work of
\citet{pease2013discussion}, who explored ways to encourage
processes of discovery ``in which chance plays a crucial role'' within
computational models of creativity.
\citet{simonton2010creative} had previously drawn relationships
between serendipity, creativity, and evolutionary processes. Of
particular interest for his analysis were generative processes which
are ``independent of the environmental conditions of the occasion of
their occurrence'' \citep{campbell1960blind}, including combinatorial as well as random
processes---a condition understood to imply teleological
``blindness.''  In a creativity setting, this condition means that one
cannot accurately predict the underlying ``fitness'' of different
ideational variants \cite[p.~159]{simonton2010creative}.  
After introducing our model and illustrating it with examples, we argue that the blindness criterion should be relaxed in line with contemporary thinking in cognitive science.

\citet{corneli2015feedback} took preliminary steps towards the system
orientation that we will develop here, and also considered how social
infrastructures might implement several of the serendipity patterns
noted by \citet{van1994anatomy}.  We are aware of recent frameworks
designed to help build systems that support the experience of
serendipity in their users \citep{niu2017framework,melo2018}: that work
testifies to the broader interest that modelling serendipity holds
within current computing research, but is different from our present aim.




We see this work as a contribution to machine discovery, a
topic that has been of interest throughout the history of AI research, was highlighted in recent computational creativity research events (ICCC'2017 panel on computational discovery),
and is increasingly relevant in contemporary applications.  \footnote{Herbert Simon contended that ``a large part of the research effort in the domain of `machine learning' is really directed at `machine discovery'{''}
\cite[p.~29]{simon1983should}.} We situate our research primarily
within the field of computational creativity.  In practical terms,
this allows us to engage at a level of abstraction above specific
implementation architectures.  After developing a model, we examine
several historical systems that illustrate the salience of the model's
features and the viability of their integration and progressive development.

The current work makes concrete contributions towards the future
development and rigorous analysis of creative systems with serendipity potential:
\begin{itemize}
	\item In this introduction, we identify the bias in the existing technical literature towards supporting serendipity in the user's experience, and propose a perspective shift from serendipity as a service to serendipity in the system. We have embraced the concept of serendipity potential in response to a classic objection to the generation of serendipity by computational means.
	\item In Section \ref{sec:literature-review}, we draw on a review of prior literature on the concept of serendipity to juxtapose existing theories and models of serendipity, in order to summarise the logical structure of serendipitous occurrences.  We understand serendipity in terms of discovery, invention and creativity, and draw connections to the associated literature to create a unified framework.
	\item In Section \ref{sec:our-model}, we synthesise a process-oriented model of systems equipped with serendipity potential which can be used to understand and qualitatively evaluate the serendipity potential of a system. We provide indicative definitions of each of six constituent phases,
\emph{perception}, \emph{attention}, \emph{interest},
\emph{explanation}, \emph{bridge}, and \emph{valuation}, based on the existing
treatment of these topics in theoretical literature.  We look as well at how people have previously approached implementation of the framework's individual components, drawing on both classic and contemporary implementation to find heuristics that can support each of the frameworks dimensions.
	\item In Section \ref{sec:system-analysis}, we provide a demonstration of our model by evaluating the serendipity potential of several documented systems developed by others.
        \item In Section \ref{sec:pursuit} we evaluate related systems developed by two of the authors, reflecting on how features of our model emerged over time.
	\item In Section \ref{sec:discussion} we discuss in turn: related work; potential directions for further use, development, and formalisation of the model; and the ways in which the model may inform future applications.
	\item In Section \ref{sec:conclusion} we put forth our
          conclusion that equipping computational systems with
          serendipity potential would be widely applicable across
          different artificial intelligence applications.  We
          emphasise that our focus is on open discovery, and that
          the model has particular relevance for future autonomous systems.
\end{itemize}

\section{The structure of serendipitous occurrences: a unified framework derived from a literature review}
\label{sec:literature-review}

To capture the intricate concept of serendipity in a model that is amenable to computational implementation, we first need a thorough understanding of the concept.  Our objective in this section is therefore to identify the factors common to existing theories of serendipity in one unified interpretation.  We will draw on related conceptualisations of \emph{creativity}, a concept that has received considerable attention in artificial intelligence research (cf.~\cite{boden1998creativity,colton2009computational,mccormack2012computers}).

At the outset it may be remarked that there are diverse perspectives on serendipity both in the theoretical literature as well as in applied work.  Usage of the term is particularly ambiguous when viewed across different computational sub-fields.  In the recommender systems context, the dominant, though not exclusive view is that serendipitous recommendations characterise items that are both surprising and valuable for the user \citep{Lu2012,Herlocker2004}.  In planning, serendipity is ``supposed to be driven by unexpected plan successes, expected but uncertain opportunities, and unexpected plan failure'' \citep{NelsonSerendipitySymp17}, as for example in the onboard planner for NASA's \emph{Deep Space One} mission \citep{muscettola1997board}. In their human-robot interaction scenario, \citet{chakraborti2015planning} consider serendipity to be ``the occurrence or resolution of facts in the world such that the future plan of an agent is rendered easier in some measurable sense.''  Here, the robot engages in planning in order to help achieve a human-sought goal.  However, this understanding appears to conflict with the typical understanding of the concept of serendipity in a scientific context, as a strictly unplanned discovery \citep{roberts}.
This diversity further motivates a return to the foundational literature.

\subsection{Etymology and selected definitions}\label{sec:etymology}
The English term ``serendipity'' derives from Horace Walpole's
interpretation of the first chapter of the 1302 poem \emph{Eight
  Paradises}---in a French translation of an intermediate Italian version of the Persian original---written by the Sufi
poet Am\={\i}r Khusrow \citep{van1994anatomy,remer1965serendipity}.  Related folktales tell similar stories \citep[p.~225]{mazur-fluke}.
The term ``serendipity'' first appears in a 1757 letter from Walpole to Horace Mann:
\begin{quote}
\emph{``This discovery is almost of that kind which I call
  serendipity, a very expressive word} \ldots \emph{You will
  understand it better by the derivation than by the definition. I
  once read a silly fairy tale, called The Three Princes of Serendip:
  as their Highness travelled, they were always making discoveries, by
  accidents \& sagacity, of things which they were not in quest
  of}[.]'' \cite[pp.~407--408]{walpole1937yale}
\end{quote}

\citet{Silver_2015} convincingly argues that Walpole appropriated
  the underlying concepts from Francis Bacon, who in turn leaned on classical Greek mythology.  Following Walpole's
coinage, ``serendipity'' was mentioned in print only 135 times over
the next 200 years, according to a survey carried out by Robert Merton
and Elinor Barber, collected in \emph{The Travels and Adventures of
Serendipity} \citep{merton}.  Merton described his own understanding
of a generalised ``serendipity pattern'' and its constituent parts:

\begin{quote}
``\emph{The serendipity pattern refers to the fairly common experience of observing an \emph{\textbf{unanticipated}}, \emph{\textbf{anomalous}} \emph{\textbf{and strategic}} datum which becomes the occasion for developing a new theory or for extending an existing theory.}''~\cite[p. 506]{merton1948bearing}~{[}emphasis in original{]}
\end{quote}
In Merton's account, the \emph{unanticipated} datum is observed while investigating some unrelated hypothesis; it is a ``fortuitous by-product'' (\emph{ibid}.). It is \emph{anomalous} because it is inconsistent with existing theory or established facts, prompting the investigator to try to unravel the inconsistency. The datum becomes \emph{strategic} when the implications of such investigations are seen to suggest new theories, or extensions of existing theories.


\citet[pp.~246--249]{roberts} records 30 entries for the term ``serendipity'' from English language dictionaries dating from 1909 to 1989.
While classic definitions required an accidental discovery, as per Walpole, this criterion was modified
or omitted later on.  Roberts gives the name \emph{pseudoserendipity} to
``sought findings'' in which a desired discovery nevertheless
follows from an accident.
\citet{Makri2012a,Makri2012b} point to a continuum between sought and
unsought findings, and highlight the role of subjectivity both in
bringing about a serendipitous outcome, and in perceiving a particular
sequence of events to be ``serendipitous.''
Many of Roberts' collected definitions treat serendipity
as a psychological attribute: a ``gift'' or ``faculty.''  Along
these lines,
Jonathan Zilberg asserts:
\begin{quote}
``\emph{Chance is an event while serendipity is a capability dependent
    on bringing separate events, causal and non-causal together
    through an interpretive experience put to strategic
    use.}''~\cite[p.~79]{zilberg2015embedded}
\end{quote}

Numerous historical examples exhibit features of serendipity and
involve interpretive frameworks that are deployed on a social rather
than on an individual scale.  For instance, between Spencer Silver's
creation of high-tack, low-adhesion glue in 1968, Arthur Fry's
invention of a sticky bookmark in 1973, and the eventual launch of the
distinctive canary yellow re-stickable notes in 1980, there were many
opportunities for Post-its\textsuperscript{\textregistered} to \emph{not} have come to
be \citep{tce-postits}.
Merton and Barber argue for integrating the
psychological and sociological perspectives on serendipity:
\begin{quote}
``\emph{For if chance favours prepared minds, it particularly favours
    those at work in microenvironments that make for unanticipated
    sociocognitive interactions between those prepared minds. These
    may be described as serendipitous sociocognitive
    microenvironments.}'' \cite[p.~259--260]{merton}
\end{quote}
Large-scale scientific and technical projects generally rely on the
convergence of interests of key actors and various other cultural factors.
For example, \cite{eco2013serendipities} describes the
historical role of serendipitous mistakes, falsehoods, and rumours in
the production of knowledge.

\subsection{Theories of serendipity and creativity} \label{sec:serendipityInvention}
Serendipity is typically discussed in the context of \emph{discovery}.
In everyday parlance, this concept is often linked with
\emph{invention} or \emph{creativity} \cite{jordanous16plos}.
However, Henri Bergson drew the following distinction:
\begin{quote}
``\emph{Discovery, or uncovering, has to do with what already exists,
    actually or virtually; it was therefore certain to happen sooner
    or later.  Invention gives being to what did not exist; it might
    never have happened.}''    \cite[p. 58]{bergson1946creative}
\end{quote}
We suggest that serendipity should be understood in terms of both
discovery and invention: that is, the \emph{discovery} of something
unexpected in the world and the \emph{invention} of an application for
the same.  Indeed, these terms provide convenient labels for the
two-part model introduced by \citet{andre2009discovery}, encompassing
the ``chance encountering of information'' followed by ``the sagacity
to derive insight from the encounter.''  \citet{mckay-serendipity}
draws on the same Bergsonian distinction to frame her argument about
the role of serendipity in artistic practice, where discovery and
invention can be seen as ongoing and diverse.  This underscores the
relationship between serendipity and creativity. At the same
  time, looking at Bergson helps to sharpen the challenge faced in any
  programmatic approach to the subject matter:
\begin{quote}
``[A] \emph{city can be constructed by photographs taken from every
  possible angle, yet this can never provide the experiential,
  intuitive value of walking in the city itself.} \ldots\ \emph{Within
  this durational context the free and intuitive action ‘drops from
  [the self] like an overripe fruit’. This drop may be seen as the
  moment of recognition within serendipity, involving a coincidence of
  prepared interior capacity with exterior conditions, in other words,
  collaboration between oneself and la dur\'ee.}''
  \cite[p.~10]{mckay-serendipity}
\end{quote}

The tension between programmatic preparedness and in-the-world action
is frequently engaged with in the computational creativity literature;
``mere generation'' is typically not deemed to be creative.  Whilst
the underlying definitions of creativity vary, two standard criteria
are variously given as ``novelty and utility,'' or ``originality and
effectiveness'' (\cite{newell:63,boden,runco2012standard}).
With a somewhat different emphasis, \citet{cropley2006praise} draws on
\citet{austin1978chase} to infuse his concept of creativity with
features of chance, and understands a creative individual to be
someone who ``stumbles upon something novel and effective when not
looking for it.''
However, Cropley questions ``whether it is a matter of luck,'' because
of the work and knowledge involved in the process of forming an
assessment of one's findings.  \citet{campbell1960blind} argues that
``all processes leading to expansions of knowledge involve a blind variation-and-selective-retention process.''  However, \citet[p.~49]{austin1978chase} remarks that:
``Nothing [suggests that] you can blunder along to a fruitful
conclusion, pushed there solely by external events.''

Cs\'ikszentmih\'alyi describes creativity similarly to
Merton's unanticipated, anomalous and strategic datum, as it arises
and develops in a social context.

\begin{quote}
``{[}C{]}\emph{reativity results from the interaction of a system
    composed of three elements: a culture that contains
   \emph{\textbf{symbolic rules}}, a person who brings
    \emph{\textbf{novelty}} into the symbolic domain, and a
    field of experts who recognize and
    \emph{\textbf{validate}} the innovation.}''
  \cite[p.~6]{csikszentmihalyi1997flow}~{[}emphasis added{]}
\end{quote}

In this case, novelty is attributed to ``a person'': even so, it is
reasonable to assume that this person's novel insights rely at least
in part on the observation of data.
Cs\'ikszentmih\'alyi's three-part model of the creative process can be compared with his
five-part phased model, comprising
\emph{preparation}, \emph{incubation}, \emph{insight},
\emph{evaluation}, and \emph{elaboration}
(\citet[pp.~79--80]{csikszentmihalyi1997flow}, adapting \citet{wallas1926art}).
\citet{Campos2002} use this later model to describe instances
of serendipitous creativity.

The more elaborate model is also a near match to the process-based model
of serendipity from \citet{lawley2008maximising}, centred on a
sequence of component-steps: \emph{prepared mind}, \emph{unexpected
  event}, \emph{recognise potential}, \emph{seize the moment},
\emph{amplify effects}, and \emph{evaluate effects}.  However, Lawley
and Tompkins's model includes a feedback loop between ``recognising
potential'' and ``evaluating effects'' that has no parallel in the
Wallas/Cs\'ikszentmih\'alyi model.  Moreover, they remark:
\begin{quotation}
``\emph{{\upshape [S]}ometimes the process involves further
    potentially serendipitous events {\upshape[a]}nd sometimes it
    further prepares the mind (at which time learning can
    {\upshape[be]} said to have taken place)}''
  \citep{lawley2008maximising}
\end{quotation}

\Citet{Makri2012a} propose a model that adapts Lawley and Tompkins,
notably by combining the ``prepared mind'' and ``unexpected event''
into one first step, a \emph{new connection}, which involves a ``mix
of unexpected circumstances and insight.''  Expanding on the notion of
a feedback loop, they suggest that a parallel process of reflection
into the ``unexpectedness of circumstances that led to the connection
and/or the role of insight in making the connection'' is important for
the subjective identification of serendipity.  Projections of value
can be updated when the new connection is exploited---for example,
when it is discussed with others.

\citet{Allen:2013:LOD:2655780.2655790} studied how the term
serendipity and its various synonyms and related terms have been used
to describe opportunistic discovery in the biomedical literature.
Three categories of usage were particularly salient:
\emph{inspiration}, \emph{mentioned findings}, and \emph{research
  focus}.  These categories of usage roughly parallel Merton's
serendipity pattern and Cs\'ikszentmih\'alyi's three-part creativity
framework.  A fourth category, \emph{systematic review}, highlighted
scholarly interest in the topic of serendipity itself.
On this note, \citet{bjorneborn2017three} surveys several theoretical
treatments beyond those mentioned above, and extracts diverse personal
and environmental factors that can promote serendipity.  We will
engage with his work later on, but for now, we have enough material to
assemble themes in line with our objective.

\def\tabularxcolumn#1{m{#1}}
\newcolumntype{Y}{>{\centering\arraybackslash}X}
\begin{table}
{\centering\small\def\arraystretch{1.2}
\begin{tabularx}{.98\textwidth}{Yc}  
\textbf{Serendipity is \raisebox{-.5ex}{$\cdots$}} & \phantom{(0)}\\[.2cm]
\end{tabularx}\offinterlineskip

\noindent\begin{tabularx}{.98\textwidth}{|Y|Y|c}  \cline{1-2}
 discovery &  invention & (1) \\
\end{tabularx}\offinterlineskip

\noindent\begin{tabularx}{.98\textwidth}{|Y|Y|c}  \cline{1-2}
{chance encountering of information} & {sagacity to derive insight} & (2) \\
\end{tabularx}\offinterlineskip

\noindent\begin{tabularx}{.98\textwidth}{|Y|Y|Y|c}
\cline{1-3}
~~~symbolic rules\newline
(that do not directly account for newly-encountered data) & novelty & {validation}& (3)
\end{tabularx}\offinterlineskip

\noindent\begin{tabularx}{.98\textwidth}{|Y|Y|Y|c} 
\cline{1-3}
findings & inspiration & research focus & (4) \\
\end{tabularx}\offinterlineskip

\noindent\begin{tabularx}{.98\textwidth}{|Y|Y|Y|Y|c} 
\cline{1-4}
unanticipated datum & anomalous datum & {strategic datum} & {new or modified theory} & (5) \\
\end{tabularx}\offinterlineskip

\noindent\begin{tabularx}{.98\textwidth}{|Y|Y|Y|Y|Y|c}  \cline{1-5}
preparation\newline
(including\newline
observations) & incubation & insight & evaluation & {elaboration} & (6) \\
\end{tabularx}\offinterlineskip

\noindent\begin{tabularx}{.98\textwidth}{|Y|Y|Y|Y|Y|Y|c}  \cline{1-6}
prepared mind & unexpected event & recognise potential & seize the moment & amplify effects & evaluate effects & (7) \\
\end{tabularx}\offinterlineskip

\noindent\begin{tabularx}{.98\textwidth}{|Y|Y|Y|Y|Y|Y|c}  \cline{1-6}
\multicolumn{2}{|p{.2787\textwidth}|}{\begin{minipage}{.2787\textwidth}
{\centering new connection

\par}
  \end{minipage}} & project value & exploit\newline connection & valuable outcome & reflect on value & (8)
\end{tabularx}\offinterlineskip

\raisebox{1cm}{\noindent\begin{tabularx}{.98\textwidth}{YYYYYYc} \Cline{1-6}{2pt}
&&&&&&\\[-.6cm]
\shifttext{1.5em}{}\newline
\textbf{\emph{perception} \mbox{of a}} \shifttext{-.6em}{\mbox{\textbf{chance event}}} &
\shifttext{1.2em}{}\newline \textbf{\emph{attention} \mbox{to salient} \mbox{detail}} &
\shifttext{1.4em}{}\newline \textbf{\shifttext{-.2em}{\mbox{\emph{focus shift}}} \mbox{achieved} \shifttext{-.5em}{\mbox{by interest}}} &
\shifttext{1.4em}{}\newline \textbf{\shifttext{-.5em}{\emph{explanation}} \mbox{of the} \mbox{event}} &
\shifttext{1.2em}{}\newline \textbf{\emph{bridge} \mbox{to a} problem} &
\shifttext{1.2em}{}\newline \textbf{\emph{valuation} \mbox{of the} \mbox{result}} & \raisebox{-.1cm}{(9)} \\ 
\end{tabularx}}

\vspace{-.5cm}
\shifttext{-2.2em}{$\underbrace{\text{\phantom{XXXXXXXXXXXXXXXXXXXXXXXXXXXXXXXXXXXXXXXXXXXXXXXX}}}$}
\vspace{.2cm}

\begin{tabularx}{.98\textwidth}{Yc} 
\textbf{All of which are operations of a \emph{prepared mind} subject to \emph{chance}.} & \phantom{(0)}\\
\end{tabularx}\offinterlineskip

\par}
\caption{Aligning ideas from several theories of serendipity and
  creativity.  Rows 1-7 show increasing detail, moving from two to six
  phases; row 8 bundles two of the steps together; row 9 summarises our
  analysis and
  provides the framework for Section \ref{sec:our-model}. Sources: (1) \citet{bergson1946creative}; (2)
 \citet{andre2009discovery}; (3) \citet{csikszentmihalyi1997flow}; (4) \citet{Allen:2013:LOD:2655780.2655790}; (5) \citet{merton1948bearing}; (6) \citet{wallas1926art} (as adapted by
  Cs\'ikszentmih\'alyi); (7) \citet{lawley2008maximising}; (8) \cite{Makri2012a}. \label{tab:theory-summary}}
\end{table}

\subsection{Distilling the literature into a framework} \label{sec:distill}

The different treatments of serendipity in many cases appear to build
on one another, and in all cases appear to be roughly aligned.
Accordingly we can distil the foregoing survey into a framework
that describes serendipitous phenomena in terms of six phases:
\emph{perception}, \emph{attention}, \emph{focus shift},
\emph{explanation}, \emph{bridge}, and \emph{valuation}.  Table
\ref{tab:theory-summary} shows graphically how we have drawn out these concepts.
In the following paragraphs, we trace through the rows of Table
\ref{tab:theory-summary} line by line, resummarising earlier
perspectives on serendipity and drawing connections between
these earlier theories and our framework.
Here we use boldface to distinguish elements of earlier theories, and
italics to distinguish elements of our framework.
\begin{enumerate}[label=(\arabic*)]
\item From \cite{bergson1946creative}: we take the notion of
  \textbf{discovery} to entail \emph{perception} and \emph{attention},
  which can potentially lead to a \emph{focus shift}.  In cases of serendipity, we understand
  \textbf{invention} to build on a discovery, through the generation
  of a novel \emph{explanation} and a \emph{bridge} to a newly
  identify problem that the explanation solves.  The solution is then
  \emph{evaluated} positively.
\item From \cite{andre2009discovery}: \textbf{chance
  encountering of information} explicitly indicates \emph{perception}
  of a chance event.  We take \emph{attention} to be implicit.  We understand the  phrase
  \textbf{sagacity to derive insight} to encapsulate what we mean by
  \emph{focus shift}, \emph{explanation}, \emph{bridge}, and \emph{valuation}.
\item From
  \cite{csikszentmihalyi1997flow}: the three-part model of creativity concerns
  interactions between a Domain, a Field, and an Individual (often
  collectively abbreviated as ``DFI'').  In cases of serendipitous
  creativity, the following occurs.
  A \emph{chance event is perceived} that cannot be fully explained
  when \emph{attended to} through the rubric of known \textbf{symbolic
    rules} which comprise a specific cultural Domain.
  A creative Individual is then inspired by the event's
  \textbf{novelty} to achieve a \emph{focus shift}, namely, to examine
  the unexplained details and generate an---\emph{a fortiori} also-novel---\emph{explanation
    of the event}.  Finally, their finding is \textbf{validated} 
  by a Field of experts when the
  explanation can be \emph{bridged} to some (new or existing) problem
  that it solves, in which case the process is deemed creative, and
  given a positive \emph{evaluation}.
\item From \cite{Allen:2013:LOD:2655780.2655790}:
  the category of \textbf{mentioned findings} suggests \emph{perception of a
    chance event} and \emph{attention to salient detail}; their
  category \textbf{inspiration} suggests a potential \emph{focus
    shift} leading to an effort to \emph{explain the event} with a
  research design that explores the serendipitous inspiration; their
  category \textbf{research focus} focuses on better understanding a
  ``fortuitous discovery'' or ``unanticipated finding'' to establish a
  \emph{bridge to a problem} that the discovery solves, towards \emph{evaluating the result}.
\item From \cite{merton1948bearing}:
  the observation of an \textbf{unanticipated} datum aligns with
  \emph{perception} of a chance event that captures our
  \emph{attention}: it is a ``fortuitous'' discovery (p. 506).
  Subsequent interest in the \textbf{anomalous} nature of the
  datum causes a \emph{focus shift} towards a \textbf{strategic}
  \emph{explanation} of the anomaly, leading to the \emph{bridge} from
  the anomalous detail to new theoretical insights.  The new (or
  extended) \textbf{theory initiated} by these investigations receives
  an at least preliminarily positive \emph{valuation}.
\item From \citet{wallas1926art}: \textbf{preparations}
  (among with we include observations) afford the
  \emph{perception of a chance event}.  Note that such preparations
  are relevant both to observing the event, and to recognising it as
  unexpected.  During a period of \textbf{incubation}, the perceiver's
  \emph{attention} may be turned towards \emph{salient details} that
  can lead to an \textbf{insight} which then leads to an
  \emph{explanation of the event}.  Here we run into some
  terminological collisions.  What we call the \emph{bridge to a
    problem} could be linked to the insight stage, but we may also
  think of it as rather close to what Wallace calls \textbf{evaluation},
  insofar as the problem that is identified at this stage is what makes
  the insight useful.  In the phase of \textbf{elaboration} (introduced
  by Cs\'ikszentmih\'alyi) the finding undergoes
  further \emph{evaluation} in new contexts.
\item From \cite{lawley2008maximising}: the
  \textbf{prepared mind} is relied upon at several stages in the
  process; indeed, as we described above, we see the prepared mind as
  vitally active throughout.  In the first instance, we can connect it
  with these authors' usage of the term ``\emph{perception}.''  As we
  noted earlier with reference to Clark's theory of predictive
  processing, the mind's previous preparations are what make the
  \textbf{unexpected event} unexpected.  Previous preparations can
  either prevent or allow \textbf{recognising potential} in a given
  observation, in part because these preparations constrain how and
  whether the individual pays \emph{attention} to the event, and
  whether or not they achieve a \emph{focus shift}. Only when the
  aforementioned steps have occurred might the person \textbf{seize the
    moment} to form a contextual \emph{explanation} of the event; and
  \textbf{amplify effects} by finding a \emph{bridge} to a problem
  that the explanation can solve.  Once all of this is done, then the
  agent may \textbf{evaluate effects}.  Note the role for a prepared
  mind in our sense---as active throughout the process---in
  supporting the ``iterative circularity'' that Lawley and Tompkins
  say may motivate several passes of recursion over the steps between
  evaluating effects and recognising potential, as well as the role of
  chance in producing opportunities to learn.
\item \citet{Makri2012a} follow Lawley and Tompkins in including
  feedback loops explicitly in their model.  Their model posits a
  \textbf{new connection} to be formed by the \emph{perception of a
    chance event} and \emph{attention to salient detail} which then
  leads the potential experiencer of serendipity to \textbf{project
    value}.  This subsequently leads to a \emph{focus shift} when
  the individual in question \textbf{exploits the new connection}. 
  We assume this is done in a somewhat \emph{explicable} or predictable way.
  Makri and Blandford assert that this
  itself is already a \textbf{valuable outcome}, i.e., it solves some
  problem directly; by \textbf{reflecting on its value} the agent may
  \emph{bridge} to a (further) problem.  An interesting aspect of the
  Makri and Blandford model is that \emph{valuation} is somewhat
  ongoing and reflecting on value may feed back into the earlier
  part of the process that projected value, leading to renewed
  interest.  As the process iterates, additional
  bridges to new problems are created, or some particular problem
  is understood in more detail.
\end{enumerate}



\subsection{Summary} \label{sec:literature-summary}

Our review of significant literature on serendipity leads us to
key features of system operation that can be described as
serendipitous.  Underpinning our analysis are foundations based on the
roles of {\em chance} and {\em the prepared mind}.  Highlights are
summarised in Table \ref{tab:theory-summary}; terms in the table are explained in the above sections.  Building on the
literature surveyed above, we describe serendipity as a form of
creativity that happens in context, on the fly, with the active
participation of a creative agent, but not entirely within that
agent's control.

While the various theories we have examined differ from one another
as to where ``insight'' takes place in the process---and some do
not mention this term---none of them seems to endorse a theory of
uninsightful serendipity.  Nevertheless,
\citet{copeland2017serendipity} has argued that ``the insight of the
individual is insufficient for bringing about a serendipitous,
scientific discovery,'' and makes a case for an understanding of
serendipity that ``goes beyond the cognitive.''  We agree with
Copeland that a contextual perspective is necessary, and we will
return to this theme in what follows: nevertheless an agent (or
\emph{agency}, per \citet{society-of-mind}) that experiences
serendipity is also necessary, and a natural place to begin 
modelling work.



\section{A computational model and evaluation framework for assessing the potential for serendipity in computational systems} \label{sec:our-model}


This section develops cognitively and computationally realistic
definitions for each of the six concepts from our synthesis of
theories in Section \ref{sec:literature-review}. 
We begin in Section \ref{sec:modelDefinition} with a high-level schematic diagram
that shows how the six phases might in principle be manifested
together in a computational system.  To demonstrate that the schematic
realistically captures common understandings of serendipity, we use it
to redescribe a famous historical case of serendipity: the invention
of PostIt\textsuperscript{\textregistered} Notes at 3M. This prepares the ground for
Section \ref{sec:modelTerms}, where we present
informally-stated but practically-inspired definitions of each of the six terms.
We support the definitions with existing foundational theories from
philosophy and cognitive science, and, for each, outline a set of
heuristics to inform future implementation work, inspired by
existing implementations.

It is important to note that each of the six phases in the model has a wide
horizon, often encompassing both good-old-fashioned AI and contemporary approaches.\footnote{For example, ``Machine Perception and Artificial Intelligence'' is the title of a book series published by World Scientific that began in 1992 and currently contains 83 volumes: \url{https://www.worldscientific.com/series/smpai}.}
We must therefore be selective rather than comprehensive in our
approach to the literature, towards 
our overall aim show that how computation might be employed to
produce serendipitous results.  Section \ref{sec:system-analysis} will
then use this model to comprehensively assess the potential for
serendipity in discrete implemented systems, particularly for computational creativity.

\subsection{A process model and rational reconstruction of a historical case study} \label{sec:ww-model}
\label{sec:modelDefinition}
\begin{figure}[h]
\begin{minipage}[b]{\textwidth}
{\centering
\resizebox{\textwidth}{!}{
\begin{tikzpicture}[
single/.style={draw, anchor=text, rectangle},
]
\node (discovery) {\textbf{\emph{Discovery:}}};

\node[single, right=8mm of discovery.east,text width=1.5cm] (poet) {\emph{generative\\ process}};
\node[single, right=6mm of poet.east] (poem) {$E$};
\draw [->] (poet.east) -- (poem.west);

\node[above left=1mm and -10mm of poet] (perception) {{\sf perception}};

\node[ellipse, draw, right=9mm of poem.east,text width=1.3cm] (critic) {\emph{feedback}};
\draw [->] (poem.east) -- (critic.west);
\node[single, above=8mm of critic.north,text width=1.4cm] (experience) {\emph{reflective\\ process}};
\node[draw,diamond,inner sep =.3mm, above right=4mm and 3mm of critic] (comment) {\raisebox{2mm}{$p\vphantom{^{\prime}}_1$}} ;
\node[draw,diamond,inner sep =.3mm, above left=4mm and 3mm of critic] (reflection) {\raisebox{2mm}{$p\vphantom{^{\prime}}_2$}} ;

\node[above right=2mm and 1mm of comment] (attention) {{\sf attention}};

\draw[->,thick] ([yshift=1mm]critic.east) to [out=0,in=270] (comment.south) ;
\draw[->,thick] (comment.north) to [out=90,in=0] (experience.east) ;
\draw[->,thick] (experience.west) to [out=180,in=90] (reflection.north) ;
\draw[->,thick] (reflection.south) to [out=270,in=140] ([yshift=1.5mm]critic.west) ;

\coordinate[below right=3mm and 7mm of critic] (mid1);

\node[single, below left=10mm and 2mm of critic] (feedback) {$T$};
\node[single, left=6mm of feedback] (selection) {$T^{\star}$};


\draw [-] (feedback.west) -- node[ fill=white, anchor=center, pos=0.5,font=\bfseries,inner sep=0pt,minimum size=1mm](selectionprocess){\guillemotleft} (selection.east);

\node[above right=3mm and -8mm of selectionprocess] (interest) {{\sf interest}};

\node[below=.65cm of discovery] (focusshift) {{\small \textbf{\emph{[Focus shift]}}}};

\draw [->] ([yshift=-1mm]critic.east) to[out=0,in=90] (mid1) to[out=270,in=0] (feedback.east);

\node[below=2.6cm of discovery] (invention) {\textbf{\emph{Invention:}}};

\node[ellipse, draw, right=12mm of invention.east,text width=1.71cm] (integrator) {\emph{verification}};

\coordinate[above left=2mm and 9mm of integrator] (mid2);

\draw [->,dashed,shorten >=5pt,>=stealth] (integrator) to[in=270] (selectionprocess);

\draw [->] (selection.west) to[out=180,in=90] (mid2) to[out=270,in=160] (integrator.west);


\node[single, below=9mm of integrator.south,text width=2cm] (explainer) {\emph{experimental\\ process}};

\node[draw,diamond,inner sep =.3mm, below right=4mm and 5mm of integrator] (question) {\raisebox{2mm}{$p^{\prime}_1$}};
\node[draw,diamond, inner sep =.3mm, below left=4mm and 5mm of integrator] (answer) {\raisebox{2mm}{$p^{\prime}_2$}};

\node[below left=1mm and 1mm of answer] (explanation) {{\sf explanation}};

\draw[->,thick] ([yshift=-1mm]integrator.east) to [out=0,in=90] (question.north) ;
\draw[->,thick] (question.south) to [out=270,in=0] (explainer.east) ;
\draw[->,thick] (explainer.west) to [out=180,in=270] (answer.south) ;
\draw[->,thick] (answer.north) to [out=90,in=200] ([xshift=1mm,yshift=-1.8mm]integrator.west) ;

\node[yshift=1mm,single, right=10mm of integrator.east] (problem) {$M$};

\draw [->] ([yshift=1mm]integrator.east) -- (problem.west);


\node[single, right=6mm of problem.east,text width=1.2cm] (pgrammer) {\emph{creative}\\ \emph{process}};
\draw [->] (problem.east) -- (pgrammer.west);

\node[above right=1mm and -10mm of pgrammer] (bridge) {{\sf bridge}};


\node[single, right=10mm of pgrammer.east] (solution) {$P$};
\draw [->] (pgrammer.east) -- (solution.west);

\node[single, below=6mm of solution.south,text width=1.6cm] (eval) {\emph{evaluation}\\ \emph{process}};
\draw [->] (solution.south) -- (eval.north);

\node[below left=1mm and -10mm of eval] (valuation) {{\sf valuation}};

\node[single, right=4mm of eval.east,text width=.3cm] (etc) {...};
\draw [->] (eval.east) -- (etc.west);
\end{tikzpicture}}

\par}
\smallskip
\end{minipage}
\caption{A boxes-and-arrows diagram, showing one possible process
  model capable of producing serendipitous results.}\label{fig:model}
\end{figure}
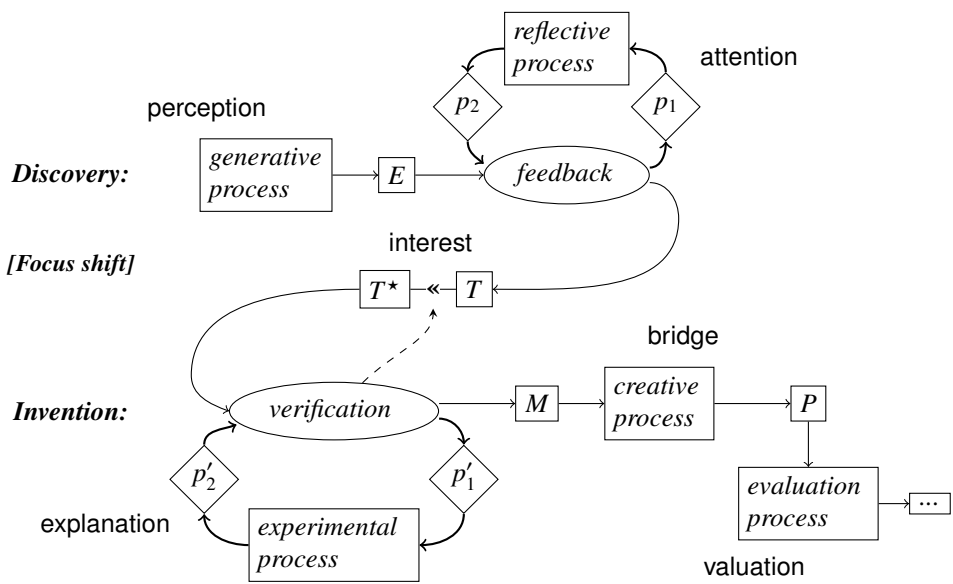

Figure \ref{fig:model} places the six phases discussed above into a
diagram outlining the idealised implementation of a (potentially)
serendipitous system.  Some steps are expanded in more detail than
others.  Other architectures might foreground different kinds
of feedback between the main steps, but to keep things simple we
have not shown all possible ways in which the process might revisit
earlier steps as it runs.
We illustrate how the diagram works in a rational reconstruction
of the invention of Post-Its\textsuperscript{\textregistered} at 3M
(quotes below are from \citet{FT3M}).
The level of detail and specificity is intermediate between the
abstract overview from the previous section and the definitions and heuristics that will be advanced in Section
\ref{sec:modelTerms}.  Before developing definitions of the individual
components, it is useful to have an example that puts
the whole process together, i.e, making the interconnections between
the phases explicit.
One immediate challenge arises in building a rational reconstruction of
the Post-Its\textsuperscript{\textregistered} example: the
story includes several steps that could informally be called
``serendipitous'' in light of the success that follows. Our reconstruction
is focused by this aim: to illustrate how a modular architecture
like the one illustrated can create serendipitous results---in this
case, using a social rather than computational infrastructure.
\paragraph{Perception of a chance event}
The first module is a \emph{generative process}.  In an
implementation, this may be based on direct observations of the
world and/or system-internal sources of chance such as a random number generator.  The output of the generative
module is understood as a chance event, $E$, that has been perceived
by the system. It is passed to the next stage.

\paragraph{\textbf{\upshape Example}}
In the 3M case study the event of interest was generated by Spencer
Silver's work in a team carrying out research on
``pressure-sensitive adhesives.''
\begin{quote}
Spencer Silver: ``\emph{As part of an experiment, I added more than
  the recommended amount of the chemical reactant that causes the
  molecules to polymerise. The result was quite astonishing. Instead
  of dissolving, the small particles that were produced dispersed in
  solvents. That was really novel and I began experimenting
  further. Eventually, I developed an adhesive that had high `tack'
  but low `peel' and was reusable.}''
\end{quote}
Here we take $E$ to include not only the bare fact of the adhesive's
creation, but also Silver's preliminary assessment.  Simply put, the
new high-tack, low-peel, adhesive would not have been created had the
reaction not captured Silver's attention and interest.  However, we
certainly cannot explain the serendipitous invention of
Post-Its\textsuperscript{\textregistered} with reference to these acts
alone.  With regard to social infrastructures, as
\citet[p.~23]{society-of-mind} remarked, ``It is not enough to explain
only what each separate agent does.  We must also understand how those
parts are interrelated---that is, how \emph{groups} of agents can
accomplish things.''

\paragraph{Attention to salient detail}
In this stage certain aspects of $E$ will be marked up as being of
potential interest, leading to $T$ in the figure.  This designation
does not in general arise all at once.  $T$ is considered to be the
result of \emph{feedback}, an abstraction over a more complex
\emph{reflective process}.  In Figure \ref{fig:model}, the reflective
process makes use of two primary functions: $p_1$ notices particular
aspects of $E$, and another, $p_2$, applies processing power and
background knowledge to enrich $E$ with additional information.  We
could call $p_1$ \emph{awareness}, and $p_2$ \emph{concentration}.
There may be several rounds of feedback applied (recursively) in order
to construct $T$.  Looking ahead to the next phase, $T$ will serve to
\emph{trigger} subsequent interest: but notice that the system is
explicitly involved in creating $T$, which does not simply arrive
wholly formed.  Nevertheless, at this stage there is little direct
evidence of how it will be used later.

\paragraph{\textbf{\upshape Example}}
In the 3M case study the key aspects of the reflective process were
implemented by Silver (who spread \emph{awareness} of the new
adhesive) together with other
employees (who developed a prototype product and gave the topic further \emph{concentration}).

\begin{quote}
Spencer Silver: ``[T]\emph{he company developed a bulletin board that
  remained permanently tacky so that notes could be stuck and
  removed. But I was frustrated. I felt my adhesive was so obviously
  unique that I began to give seminars throughout 3M in the hope I
  would spark an idea among its product developers.}''
\end{quote}

\begin{quote}
Art Fry: ``\emph{I was at the second hole on the golf course, talking
  to the fellow next to me from the research department when he told
  me about Spencer Silver, a chemist who had developed an interesting
  adhesive. I decided to go to one of Spencer's seminars to learn
  more. I worked in the Tape Division Lab, where my job was to
  identify new products and build those ideas into businesses. I
  listened to the seminar and filed it away in my head.}''
\end{quote}

\paragraph{Focus shift achieved through interest}
The trigger $T$ thus consists of the original event, $E$, together
with a range of newly-added metadata and markup.  A focus shift
selects ({\bf \guillemotleft}) some elements from this complex object,
potentially using them to retrieve additional data.  The result is
``of interest,'' denoted above by $T^{\star}$.

\paragraph{\textbf{\upshape Example}}
In the 3M case study, the information that Fry had filed away
before ($T$) became interesting when he realised that he ``had a
[related] practical problem'' ($T^{\star}$).

\begin{quote}
Art Fry: ``\emph{I used to sing in a church choir and my bookmark
  would always fall out, making me lose my place.  I needed one that
  would stick but not so hard that it would damage the book.  The next
  morning, I went to find Spencer and got a sample of his adhesive.}''
\end{quote}
In this case, the adhesive becomes interesting insofar as it could
potentially used to create a re-stickable bookmark.  3M allowed its
employees to selectively allocate 15\% of their time
\citep{tce-postits}, and Fry decides to initiate his own experiments.

\paragraph{Explanation of the event}
The now-interesting trigger $T^{\star}$ is submitted for
\emph{verification}, which Figure \ref{fig:model} depicts as an
abstraction over an \emph{experimental process}, whose operations here
again consist of two primary functions: \emph{theory generation},
$p^{\prime}_1$, and \emph{theory checking}, $p^{\prime}_2$.  The
result of this process is a \emph{model}, $M$.  The dashed arrow in
the diagram is meant to indicate that the focus shift stage may be
revisited and new selections made as this process progresses, i.e.,
$T$ may become interesting for new or different reasons as the
experimental process progresses.

\paragraph{\textbf{\upshape Example}}
In the 3M case study, Fry already has in mind the theory
($p_1^\prime$) that re-stickable bookmarks can be made using the new
adhesive.  Fry creates and adjusts a working prototype  ($p_2^\prime$)
on the way to verifying his theory.

\begin{quote}
Art Fry: ``\emph{I made a bookmark and tried it out at choir practice; it
didn't tear the pages but it left behind some adhesive. I needed to
find a way to keep the particles of the adhesive anchored to the
bookmark.  After a few experiments, I made a bookmark that didn’t
leave residue and tested it out on people in the company.}''
\end{quote}
Note that in this case the event $E$ has not been explained in terms
of ``how'' but rather, contextually, in terms of ``so what?''  The
nature of the explanation will differ from case to case.  The common
feature is the creation of a causal model of some sort.  In this case,
the causal model $M$ is a \emph{method} for creating re-stickable
bookmarks that don't leave residue.

\paragraph{Bridge to a problem}
Here the system forms a connection (``bridge'') between the
explanation in the form $M$ and some as-yet-unspecified problem, $P$.
The schematic represents this step in one block, a \emph{creative
  process}.  This is clearly underspecified, but we shall describe
different possible implementation strategies shortly, in Section
\ref{sec:modelTerms}.

\paragraph{\textbf{\upshape Example}}
Let's see how this process worked in the 3M case study.  Fry now had a
prototype, but so far it didn't solve a very interesting problem.
(``They liked the product, but they weren’t using them up very
fast.'')  But then:

\begin{quote}
Art Fry: ``[O]\emph{ne day, I was writing a report and I cut out a bit
  of bookmark, wrote a question on it and stuck it on the front. My
  supervisor wrote his answer on the same paper, stuck it back on the
  front, and returned it to me. It was a eureka, head-flapping moment
  -- I can still feel the excitement.  I had my product: a sticky
  note.}''
\end{quote}

It would seem that no one, including Fry, had thought about this
problem before: how can we easily make notes on a document, without
marking up the document itself, and without introducing other separate
sheets of paper that would need to be stapled or paper-clipped to the
document, or that might get lost?

Indeed, without knowing the solution in advance, or having $M$ in mind
and re-stickable bookmarks to hand, the problem might even sound like
a contradiction in terms.  It would probably have been impossible to
solve it very well using conventional methods
\cite[p.~90]{altshuller2007innovation}.  But remember that Fry was
part of the Tape Division.  By cutting off a piece of the bookmark,
and affixing it to the front of the report, he was using the bookmark
like one might have used a piece of tape---which would have been
another semi-conventional solution, different from staples and
paperclips, for affixing a separate sheet of paper.  However, the new
``sticky note'' had several advantages over tape: it could be written
on directly and easily removed later.  Thus, we may rationally
reconstruct the bridge to $P$ via an intermediate virtual solution of a note
taped to the report's cover.

\paragraph{Valuation of the result}
The new problem, $P$, which now conveniently has a solution in the
form of $M$, is passed to an \emph{evaluation process}, and, from
there, to further applications.  One possible class of applications
would be a change to any of the modules that participated in the
workflow, corresponding to the potential for learning from
serendipitous events noted by \citet{lawley2008maximising}.

\paragraph{\textbf{\upshape Example}}
The 3M example shows that evaluation can itself be a complex process:

\begin{quote}
Art Fry: ``\emph{We made samples to test out on the company and the
  results were dramatic.  We had executives walking through knee-deep
  snow to get a replacement pad.  It was going to be bigger than Magic
  Tape, my division’s biggest seller.  In 1977, we launched Post-it
  Notes in four cities.  The results were disappointing and we
  realised we needed samples.  People had to see how useful they
  were. Our first samples were given out in Boise, Idaho and feedback
  was 95 per cent intent to re-purchase.  The Post-it Note was
  born.}''
\end{quote}
Notice that in this case the approach to valuation is itself updated on the fly.

\subsection{Definitions of the model's component terms} \label{sec:modelTerms}

We now present short definitions of each component, which we support
with references to foundational literature from cognitive science and
philosophy, as well as heuristics that relate to the current status of
implementation work as evidenced by computing literature.  Our
thinking in this section is informed by the ``predictive processing''
framework advocated for example by \citet{friston2009free}, \citet{clark2013whatever}, and others.  A central idea in such
theories is that perceived events are only passed forward to higher
cognitive layers if they do not conform with our prior expectations.
This perspective highlights the fact that, going beyond Pasteur's famous idiom, chance
not only \emph{favours}, but also \emph{shapes} the prepared mind.
Thus, for example, \citet[p.~137]{boden} notes that ``neural networks
learn to associate (combine) patterns without being explicitly
programmed in respect of those patterns.''

Multi-level architectures
abound in AI; one example comes from \citet{singh2005architecture}, where the
first level beyond ``innate reactions'' is ``learned reactions'';
higher levels include ``deliberative thinking'', ``reflective
thinking'', ``self-reflective thinking'' and ``self-conscious
thinking.''  \citet{sloman2002framework} place somewhat similar concepts
in a two-dimensional schema which they suggest can
be used to compare different architectures.
While sharing concepts of hierarchical control with such models,
theories based on predictive processing
``upend'' the classical input/output paradigm, recentring on thermodynamic energy
transfer: their models of control are continuous and ``there are no disconnected
moments of perception of the world, since the world wholly envelops the
agent throughout its lifespan'' \cite[pp.~9--10]{10.3389/frobt.2018.00021}.
\citet{kockelman2011biosemiosis} develops a related line of thinking
from a semiotic perspective, pointing out that processes of
``sieving'' and ``selection'' are not just properties of the mind but
also of the environment.
Upon considering these reflections, we cannot subscribe to the view that
serendipity is ``a process of discovering with a completely open
mind'' \citep{darbellay2014interdisciplinary}.  The mind will in
general have been shaped by previous interactions with the world.

Furthermore, while we necessarily must present the phases of our model in order, we
hereby make explicit the assumption that phases encountered earlier
can be returned to from temporally-later ones.  Because the phases
build on one another, we propose that they must be encountered in
temporal order, backward-directed moves notwithstanding.  In other
words, we allow the process to jump backward, and only jump
forward to steps that have been encountered already.  This
does not imply that future steps are always entirely impossible to
anticipate, however.  Thus, for example, Pasteur's research has been described
as ``use-inspired'' \citep{stokes1997pasteur}.  Some famous pseudoserendipitous
discoveries, such as the treatment of disease with safe antibiotics,
were pursued in broad outline
long before the details became clear \citep{fleming}.

In this respect we note that Friston's model of predictive processing
makes more specific and detailed assumptions about structure and
interconnection than we will adhere to here, namely that ``error-units
receive messages from the states in the same level and the level
above; whereas state-units are driven by error-units in the same level
and the level below'' \cite[p.~297]{friston2009free}.  In simpler
biologically-inspired terms, ``the brain generates top-down
predictions that are matched bottom-up with sensory information''
\cite[p.~2]{Bruineberg2018}.  The mismatch between sense data and existing
ubiquitously generative models is how prediction errors are said to arise, which the system then strives to correct.
Here, the simpler account of interconnections between the modules that we developed
in Section \ref{sec:ww-model} guides our work.  Our model also has integral generative
aspects, but they differ at the different phases.

To emphasise, our intention in this section is to give a plausible
general account of the six phases from which our model is comprised:
we offer a top-down analysis.
Accordingly, we do not give exhaustive technical definitions, nor do
we make detailed assumptions about the overall architecture.  The
heuristics are intended to present practical advice that could be used to
increase a system's serendipity potential with respect to each phase.

\paragraph{\textbf{\upshape Primitive Notions}}
We cast our definitions in terms of the primitive notions of
``system'', ``event'', ``context,'' and ``object'' which for
conciseness we describe by way of examples.
\begin{itemize}[label=--]
\item By {\em system} here mean a computational system.  Extant
  examples include systems which generate music, art or poetry; make
  discoveries in drug design, generate scientific discoveries or prove
  mathematical theorems; carry out predictive modelling such as
  classifying an email as spam or not; personal conversational
  assistants; systems which play games such as Go or Chess, and so on.
  The system has the ability to process data and make evaluative
  decisions.
\item By {\em event} we mean some form of input or generated
  data. Examples could include partial fragments of music, art or
  poetry which have been input or generated; external events such as
  new data in a dynamic world, a new classification weighting, an
  email to classify, a conversational turn, a piece of information
  about the weather, a Chess move, and so on.  
\item By {\em context} we mean a specific set of events, data,
  algorithms, generative and evaluative mechanisms, search strategies,
  etc., that a system can access, and which may be related to a
  current goal or problem. An example of a context is a particular
  sequence of questions asked to a conversational agent, possible answers and
  their sources, a way of ranking possible answers and any other
  information or techniques necessary to produce an answer.
\item An \emph{object} is an element of a context.
\end{itemize}

\begin{defn}
\label{def:perception}
\hypertarget{def:perception}{}\textbf{Perception:} The processing of
events that arise at least partially as the result of factors
outside of the system's control.
\end{defn}

\setlist[description]{font=\normalfont\itshape,itemsep=0pt,style=unboxed,leftmargin=*}

\paragraph{\textbf{\upshape Foundations}}
~
\newline\emph{System-environment relationships differ widely, and develop differently.}
The environment may be more or less observable; events
  may appear to be more deterministic or more stochastic in nature
  \cite[pp.~42--44]{russel2003artificial}. The system may be able to
  self-program using the environment, possibly via 
  interaction with other systems \cite[esp.~p.~234]{clark1998being}.
  The system's perceptual features and limitations can vary with time,
  location, the state of development of the system, and other factors.
\vspace{.1cm}\newline\emph{Chance can play various roles in shaping perception.} For
  \citet[p.~99]{hume1904enquiry} \emph{chance} denotes the absence of
  an explanation; for \citet{peirce1931necessity} it is one of several
  fundamental aspects of reality; for 
  \citet[p.~234]{bergson1983creative}, it ``objectifies the state of
  mind'' of one whose expectations are confounded.  Unexpected
  events constitute novel perceptions, and can motivate action that leads to further perceptions.
\vspace{.1cm}\newline\emph{The system has limited control.} The world is not entirely under
  the control of the system:
  furthermore,
  perceptions necessarily constitute an incomplete picture of reality
  \cite{hoffman2015interface}.  As in Figure \ref{fig:model} and its
  accompanying discussion, our model allows events to arise through generative
  methods, but this again implies a circumscribed locus of control,
  namely, over the generative process, but necessarily over the results.\footnote{For example,
  there is a difference between generating elements in a sequence of 1's, the
  elements of which are predictable \emph{a priori}, and generating
  additional digits of the decimal expansion of $\pi$, which
  should be replicable \emph{a posteriori} but which in practice involves nontrivial computation.}
Taking a view grounded in predictive processing, \citet[pp.~2, 17--18]{10.3389/frobt.2018.00021} emphasise 
the epistemic and existential salience of generative models
(and continuous action/perception loops, including proprio- and intero-ception)
for both organisms and future robots.  The basic view is
that 
``we harvest sensory signals that we can predict'' \citep{friston2009free},
though such predictions are fallible.

\paragraph{\textbf{\upshape Heuristics}}
~
\newline\emph{To create the possibility for varied patterns of inference to arise, support rich interfaces.} 
Computer support for natural
  language interaction remains limited. Human-Computer Interaction
  researchers have experimented with a wide range of alternative interface
  designs (e.g., ranging from 
head tracking and gesture tracking \citep{turk2000perceptive} to
interaction through dance \citep{jacob2015viewpoints}
  and with physical models \citep{stopher2017technology}).
\vspace{.1cm}\newline\emph{To reduce constraints, allow features to be defined inductively.}
Rather than building systems that simply notice
  pre-conceived features of the environment, recent research has dealt
  with systems that independently discover perceptible features
  \citep{inceptionism}.
\vspace{.1cm}\newline\emph{Organise and process perceptions differently depending on the tasks undertaken.}
 Humans have \emph{head direction} and \emph{grid cells} that
  help define our relationship to the environment, and that support
  spatial navigation tasks.  Similar phenomena have been reproduced in
  machine learning programs \citep{Banino2018,cueva2018emergence}.
  However, AI systems often operate in environments that are
  structured very differently from their human analogues, e.g., when
  machine learning is applied to text corpora.  Rather than adjusting the
  underlying source of perceptions, it may be be preferable to build constraints on action
  that give an ``explicit characterization of acceptable behavior''
  \citep[p.~356]{caliskan2017semantics} within the environment.

\begin{defn}
\label{def:attention}
\hypertarget{def:attention}{}\textbf{Attention:} The system's directed
processing power applied to a perceived event, which is accompanied by an
initial evaluation.
\end{defn}

\paragraph{\textbf{\upshape Foundations}}
~
\newline\emph{Adaptive attention is related to surprise.} According to
  \citet{clark2013whatever}, an event only draws attention when the
  perceiving agent did not anticipate it.
\vspace{.1cm}\newline\emph{Learning, context, and meaning begin to arise together with attention.}
  ``Punctuating events'' \cite[p.~301]{bateson-logical-categories}
  from a stream of data is a basic form of attention.  Identifying
  patterns that are stable over time, which then begin give the data
  ``context and interpretation'' \citep{rowley2007wisdom} is another.
\vspace{.1cm}\newline\emph{To some approximation, features of the environment will be attended to.} This is a version of the hypothesis
  that hierarchical structures in the environment will be mirrored by
  \emph{adaptive} agents (\cite{simon1962architecture,simon1995near}).
  Outside intervention may be needed to optimise learning about tasks
  with complicated problem/subproblem structure
  \citep{goldenberg2004may}.

\paragraph{\textbf{\upshape Heuristics}}
~
\newline\emph{Attention can be understood as competition for scarce processing resources.}
   For example, visual attention has been
  described this way \citep{helgason2012attention}, and parallels can
  be seen in grammar-inducing processes \citep{wolff1988learning}.
  Taken as a metaphor, this extends to ``the mental grammar of the
  investigator'' and the way they ``parse their conceptual domain''
  \citep{doi:10.1080/10400410409534554}.
\vspace{.1cm}\newline\emph{Attention can be time-delineated.}  In his design of the
  discovery system {\sf AM}, Doug Lenat assigned ``a small
  interestingness bonus'' \cite[p.~281]{lenat1984and} to each new
  concept the system created.  The bonus decayed rapidly with each new
  task undertaken, but in the mean time, it made the new concept more
  likely to be used.  This was inspired by a similar but more complex
  ``Focus of Attention'' facility in the blackboard system {\sf
    Hearsay-II} \citep{lesser1977retrospective}.
\vspace{.1cm}\newline\emph{Competition may be less natural when we can take advantage of parallelism.}  Humans have the ability to process complex activities
  in parallel \cite[pp.~40--42]{blackmore2005consciousness}; as we saw
  in Section \ref{sec:ww-model}, social infrastructures can distribute
  features of attention, such as awareness and concentration, across a population.
  \emph{Joint attention} is one such important social phenomenon.  In
  related computational work \citet{zhuang2017parallel} describe a
  system for parallel attention that recurrently identifies objects in
  images.  It makes use of both image-level attention and text-based
  proposals (the latter directed to image regions), allowing image
  contents to be identified in a dialogue format.  \citet{xu2015show}
  also worked on image captioning, this time using a long short-term memory
  (LSTM) network that independently selected image regions.
  LSTMs are detailed computational models of neurons that are
  capable of learning long-term dependencies.  Xu et al trained their networks
  using models of ``soft'' and ``hard'' attention: the latter did somewhat better for the metrics
  considered.  For a navigation task, \citet{vemula2017social} had success using ``soft
  attention over all humans in the crowd,'' 
  i.e., not simply those people who are nearest.

\begin{defn}
\label{def:focus-shift}
\hypertarget{def:focus-shift}{}\textbf{Focus Shift:} A reassessment in
which an object that had been given a neutral, or even negative value,
becomes more interesting.  This may
happen, for instance, if a change of context means that a previously
encountered object is now considered to be relevant. 
\end{defn}

\noindent Given the central nature of the focus shift in our
  model, we expand its preconditions in more detail.

\begin{customdefn}{3A}\label{def:focus-shift-ability}
\hypertarget{def:focus-shift-ability}{}\textbf{Ability to Focus
  Shift:} Let $E(o,c)$ be the evaluation performed by the system
according to a set of evaluation criteria of a given object $o$ in a
given context $c$. A focus shift occurs when, for object $o$ and
context $c_1$, $E(o,c_1) \leq \theta$ for a given threshold $\theta$;
and the system either:
\begin{enumerate}[label=\emph{(\roman*)},leftmargin=2cm]
\item retrieves an existing context $c_2$ such that $E(o,c_2) >
  \theta$
\item generates a new context $c_2$ such that $E(o,c_2) > \theta$,
  or
\item changes its evaluation criteria to $E'$ such that $E'(o,c_1) >
  \theta$
\end{enumerate}
(Or some combination of {\em (i) - (iii)}.)  A system that can perform
one or more of these operations has \emph{the ability to focus shift}.
\end{customdefn}


\paragraph{\textbf{\upshape Foundations}}
~                                         
\newline\emph{Assess the data's potential for strategic usefulness.} 
  In evolutionary computing, 
  \emph{fitness} is typically an attribute of an agent, often modelled as a
  scalar value.  Now, instead, we might understand the agent's objective
  functions to give rise to a fitness landscape that can
  drive transformation of the data the system encounters, or
  cause it to be cast aside.
  Simonton makes use of a somewhat related concept of fitness,
  distinguishing between \emph{blind} and \emph{sighted} selection
  \cite[p.~159]{simonton2010creative}: he proposes a fitness measure  for
  selected items which is understood as a
  measure of their utility for the agent (which is what what the agent
  or may not may be blind to).
  Definition \ref{def:focus-shift-ability} makes no assumptions about the
  actual utility of selected items.
\vspace{.1cm}\newline\emph{Interest is related to curiosity.}
Berlyne distinguished between \emph{perceptual} and \emph{epistemic}
curiosity, while positing a relationship between them: one ``leads to
increased perception of stimuli'' and the other to ``knowledge''
\cite[p.~180]{berlyne1954theory}.  He posited that responses would be
strongest in an ``intermediate state of familiarity'' which triggered
conflict, whereas ``too much familiarity will have removed conflict by
making the particular combination an expected one'' (p.~189).
Accordingly, such curiosity depends on prior preparations.
In some reinforcement learning models, a \emph{novelty bonus}
``acts like a surrogate reward'' and ``distorts the landscape
of predictions and actions, as states predictive of future novelty
come to be treated as if they are rewarding'' \cite[p.~554]{kakade2002dopamine}.
Whether or not novelty is interesting in and
of itself,  the system's initial assessment
motivates it to look for further information or ``new connections,''
as per \citet{Makri2012a}.  This effort is expected to
yield a future payoff, whether in terms of additional novelty,
more efficient organisation of the system's knowledge base,
or in some other way.  Crucially, interest is not related exclusively to curiosity, but to a whole set of intrinsic motivations.
\vspace{.1cm}\newline\emph{Context change is a possible basis for belief revision.}
  \citet{logan1994modelling} use the notion of \emph{belief revision}
  to model situations of collaborative information-seeking.  Ground
  assumptions are shared in the context of such dialogues, and can
  change as conversations progress.  In our model, the focus shift
  similarly causes the context to change, so that the ground
  assumptions, including ways of evaluating data, are no longer the
  same.  \citet{harman1986change}
  treated the implications of changing circumstances for bringing
  about a ``reasoned change of view'' (p.~3); he described
  previous work by \citet{Doyle:1980:MDA:889488} 
  on the system {\sf SEAN}, which incorporated defeasible reasoning,
  as one of only a few earlier efforts in this area.
  More recently, \citet{clarke2017assertion} argues that \emph{belief} is
  context-sensitive, depending for example on purpose, and on the
  stakes involved.  Thus, in a dialogue, the sincerety of a
  given remark is linked to the context, not just to the remark's
  propositional content.  
                                         
\paragraph{\textbf{\upshape Heuristics}} 
~                                         
\newline\emph{Interest can be linked to novelty in order to inspire learning.}
In the case of
  Velcro\textsuperscript{\texttrademark}, the focus shift occurred in
  quite a literal fashion, when de Mestral examined burrs under a
  microscope.  This example provides another useful mnemonic: burrs'
  hooks allow them to ``hitchhike'' into new contexts
  \cite[\textsection1.1]{jenkins2011bio}.  
  \citet{patalano1993predictive} describe the related mental phenomenon of
  \emph{predictive encodings} that record ``blocked goals in memory in such a
  way that they will be recalled by conditions favorable for their
  solution.''
  The {\sf Curious Design Agents} developed by 
  \citet{Saunders2007} evolve artworks in respect to a sophisticated
  measure of interestingness.  These agents cluster artworks together,
  and assess the novelty of new inputs by means of classification
  error.  They then determine a new artwork's interestingness by
  mapping its novelty to an inverse-U-shaped curve, inspired by the
  Wundt curve (cf.~\citet[pp.~17--19]{berlyne2013pleasure}).  This
  model is useful for ``modelling autonomous creative behaviour'' and
  can ``promote life-long learning in novel environments''
  \citep{saunders2010curious}.
  A similar conception of interest is has been applied to ``generate
  art with increased levels of arousal potential in a constrained way
  without activating the aversion system,'' using a variant of
  Generative Adversarial Networks to motivate the creation of
  visual artworks that exhibit ``stylistic ambiguity'' \cite[p.~97]{elgammal2017can}.
  Mathematicians, such as
  \citet{birkhoff1933aesthetic}, have proposed many mathematical
  theories of aesthetics, though philosophers have just as often
  refuted them \cite[p.~4]{hyman2006objective}.  In J\"urgen
  Schmidhuber's work, interestingness is positioned as the ``first
  derivative of subjective beauty'' \citep{schmidhuber2009art}---where
  beauty is understood as \emph{compressibility}.  Here, phenomena that
  maximise prediction error drive curiosity.
  \citet{javaheri2016analysis} apply related measures of
  \emph{information gain} and \emph{Komolgorov complexity} to evaluate
  and drive the evolution of 2D patterns generated by cellular
  automata.
\vspace{.1cm}\newline\emph{Interest can be linked to aesthetics in order to capture varied notions of fitness.}
  \citet{dhar2011high} describe an ``aesthetics
  classifier'' that can determine the potential interestingness of
  images in terms of \emph{high level content} and \emph{compositional
    attributes} such as ``people present'', ``opposing colors'', and
  ``follows rule of thirds.''
  \citet{DBLP:journals/corr/abs-1802-10240} applied machine learning
  to a corpus of digial photographs with ratings and reviews, and
  generated new textual descriptions and rating predictions based on
  the crowdsourced descriptors.  {\sf DARCI} (short for Digital ARtist
  Communicating Intention) is a generative program which similarly
  links crowdsourced image descriptions to extracted features
  \citep{norton2013finding}.  It evolves input images using a fitness
  function that optimises for a combination of \emph{appreciation},
  defined in terms of describability, and \emph{interest}, which is,
  as above, an inverse-U-shaped measure of similarity to the input
  image.
\vspace{.1cm}\newline\emph{Beauty is in the eye of the beholder.} \citet{corneli2016x575}
  follow \citet{waugh1980poetic} in describing \emph{complexity} and
  \emph{coherence} as two key aspects of poetic beauty.  With regard
  to their implemented system that generates linked verse: ``A reader
  may identify some fortuitous resonances [in the system-generated
    poems] but the system itself does not yet recognise these
  features.''  \citet{veale2015game} discusses a related \emph{placebo
    effect} among readers of computer-generated tweets, and the
  broader role that ``an active and receptive mind'' plays in our
  interactions with the world.

\begin{defn}
\label{def:explanation}
\hypertarget{def:explanation}{}\textbf{Explanation:} This is a model that predicts functional, operational, or statistical behaviours that relate the previously-unexpected event to its newly-retrieved context.
\end{defn}
\paragraph{\textbf{\upshape Foundations}}
~                                         
\newline\emph{A new model yields an improved ability to make a prediction.}  Our
  assumptions about chance, described earlier, insist that
  the perceiving agent has at best a limited ability to predict the events it perceives.
  The explanation stage now enables the agent to make predictions
  \cite[p.~389]{sowa2000knowledge}.  Explanatory success depends on
  the system's skills, and both prior and new knowledge.  However,
  these explanations are again limited.
  \citet[p.~101]{swirski2000between} points out that to be effective,
  explanation needs ``a stopping rule''---for example, ``the standard causal
  pattern in the social sciences'' requires only ``a description of
  the actions and the motivations behind them that were sufficient to
  produce a change in the circumstances.''
\vspace{.1cm}\newline\emph{There are different kinds of viable explanations.} In the 3M
  example, the explanation focused on ``so what,'' i.e., on showing
  that the new adhesive could be used to make re-stickable bookmarks, and ultimately, a saleable product.
  However, explanations need not focus on outcomes.
  An explanation can be related purely to ``how.''  For instance,
  van Andel describes the example of Simcha Blass, who
  \begin{quote}
    ``\ldots\ \emph{happened to pass a row of trees. He
    noticed that one of the trees was much taller than the others. On
    investigation he found that, although the soil around the tree was
    dry, water was continually dripping from a nearby leaking
    connection in a water pipe.}'' \cite[p.~640]{van1994anatomy}
  \end{quote}
  This
  is a fine ``how'' explanation: the practical usefulness of Blass's model arose
  only later.  According to Aristotle, the fundamental question that
  must be addressed is ``why?''  \cite{sep-aristotle-causality}:
  answers are to be demonstrated in terms of ``principles and causes''
  \cite[Book Gamma, p.~81]{lawson1998metaphysics}.
  But crucially, even an incorrect explanation could turn out to be useful
  later on: ``reliable'' explanations are not always correct, or may
  only be correct within circumscribed regimes.
\vspace{.1cm}\newline\emph{The system creates an explanation of the event for itself.}  At
  this stage the system is not, in general, aiming to explain its
  behaviour to someone else, or otherwise make its behaviour
  transparent (in the sense of \emph{Explainable AI}
  \cite{lane2005explainable}).  Nevertheless we may think of
  explanation as an expository device or ``framing''
  \citep{pease2011computational} that relies on the system's ability to
  retrieve a suitable context, and to establish relationships between
  elements of this wider context.  Explanatory prowess is not simply a
  matter of paying attention, but depends in particular on having
  learned ``what to pay attention to'' \cite[p.~4]{levin1975bateson}.
  Notice that requirements arising in this stage can push back
  on earlier stages. ``[T]he methods and assumptions on which a
  systematic investigation is built selectively focus the researcher's
  attention'' \cite[p.~131]{floppyearedrabbits1958barber}.
                                         
\paragraph{\textbf{\upshape Heuristics}} 
~                                         
\newline\emph{Experiments can have limited scope and still be useful.}  For
  example, \citet{delamaza1994generate} describes two implementations
  of a ``Generate, Test, and Explain'' architecture.  The programs
  involved used decision trees to connect secondary contextual
  information (e.g., macroeconomic indicators) to more elementary
  data-driven predictions (e.g., of stock market behaviour).  The aim of this work was solely
  to ``connect the `correlations' uncovered by the generate and test
  module to the causal model provided by the domain theory''
  (\emph{ibid.}, p.~50).  A strategic use for these connections could in principle be
  found later.
  The system {\sf
    KEKADA}  by \cite{kulkarni1988processes} is cited by de la Maza as an example of a system that can
  directly refine the domain theory.
\vspace{.1cm}\newline\emph{Given a sufficiently rich background, only a small amount of new data is needed.}
  The term \emph{explanation-based learning}
  \citep{ellman1989explanation,cohen1992abductive} denotes a process in
  which an explanation of one event leads to a rule that can be
  applied to similar events in the future.  This typically requires
  significant background knowledge.  Imitation learning, learning from
  demonstrations, learning by example, and one-shot learning are
  related concepts (see, for example, \cite{cypher1993watch}).
  \emph{Case-based reasoning} formulates background knowledge as an
  extensive catalogue of somewhat-similar ``cases'': here explanation
  may play a role in determining how two cases match
  \cite[p.~11]{aamodt1994case}.
\vspace{.1cm}\newline\emph{Learning is less efficient, but more widely applicable, than knowing.}
  The system {\sf Hacker}, created by \citet{sussman1973computational}, was
  able to ``diagnose five classes of mistake and adapt differentially
  to them, generalizing its adaptive insights so that they can be
  applied to many problems of the same structural form''
  \citep{boden1984failure}.  However,
  \begin{quote}
    ``\emph{Hacker is not as good at
  solving blocks world problems as would be a much simpler program
  that just goes about it directly with some good heuristics and a
  minimum of exploration.  Hacker's justification is as an
  epistemological model, not as a real problem solver}''
  \citep[p.~17]{levin1975bateson}.
  \end{quote}
  Sussman-style ``critics''---which find,
  fix, and in future avoid bugs---have been widely used in the planning literature
  \citep{Sacerdoti:1975:SPB:907010,Young1994,erol1995critical,singh2005alternate,kaelbling2011hierarchical}.
  For example, critics have helped create video game
  characters that make situationally-appropriate plans in complex,
  changing, environments \citep{hawes2001anytime}.
\vspace{.1cm}\newline\emph{Communication between agents can transfer causal information.}
  \citet{moore1995participating} and \citet{cawsey1992explanation}
  describe systems that provide explanations to the user in
  interactive dialogues.  Subsequent research compared
  ``mixed-initiative'' and ``non-mixed-initiative'' dialogues using
  computer simulations \citep{ishizaki1999exploring}.  There
  are other ways to share and integrate causal information when it has
  formal representations \citep{GeiHofSch16}.  As is well known from
  research on social dilemmas, thin communication protocols constrain
  agents' ability to cooperate; however, sufficiently complex agents
  can learn to cooperate even with limited communication bandwidth
  \citep{leibo2017multi}.

\begin{defn}
\label{def:bridge}
\hypertarget{def:bridge}{}\textbf{Bridge:}
The path and set of mechanisms used to transform the
triggering event and output from subsequent processing steps into a problem to solve.
Mechanisms often include reasoning
techniques, such as abductive inference or analogical reasoning, and
may rely on new social arrangements or physical prototypes.  The bridging process may
have many steps, and may feature chance elements.
\end{defn}

\paragraph{\textbf{\upshape Foundations}}
~                                         
\newline\emph{It is sometimes necessary or desirable to go beyond explanation.}
  The bridging process can be
   outlined by comparing a positive example with a
  corresponding counterexample.  Nearly 60 years before Fleming,
  Eugene Semmer both discovered and also cursorily explained the
  curious effects of \emph{penicillium notatum}---but he did not find
  a bridge to the vital problem his discovery could have solved
  \cite[p.~75]{cropley2013creativity}.  His ``methods and
  assumptions'' \cite[p.~131]{floppyearedrabbits1958barber}
  constrained his thinking.
\vspace{.1cm}\newline\emph{Two cases: pseudoserendipity versus true serendipity.}  The ``eureka'' or ``aha'' moment has been
  modelled computationally by \citet{thagard2011aha} using a form of
  concept blending.  These authors assert that ``human creativity
  requires the combination of previously unconnected mental
  representations constituted by patterns of neural activity'' (p.~1).
  The notion of a \emph{bridge} is suggested, but such a connection
  may be a sought finding.   The Bergsonian distinction treated in
  Section \ref{sec:literature-review} emphasises making a connection not
  simply between representations, but to a novel problem:
  ``[originally] stating the problem is not simply uncovering, it is
  inventing'' \cite[p. 58]{bergson1946creative}.  Due to its novelty,
  an original problem cannot be fully known in advance, though the investigator
  may invent such a problem whilst in quest of something else.
  \citet[p.~3]{Figueiredo2001} made the distinction between
  serendipity and pseudoserendipity particularly crisp by introducing
  the ``serendipity equations'':
\begin{center}
\begin{tabular}{c}
  \emph{pseudoserendipity}\\[.1cm]
$\begin{array}{c}
P1 \subset (\mathit{KP}1)\\
M \subset (\mathit{KM})
\end{array} \Rightarrow S\hspace{-.02em}1 \subset (\mathit{KP}1, \mathit{KM}, \mathit{KN})$
\\[.4cm]
\emph{serendipity}\\[.1cm]
$\begin{array}{c}
P1 \subset (\mathit{KP}1)\\
M \subset (\mathit{KM})
\end{array} \Rightarrow
\begin{array}{c}
P\hspace{.02em}2 \subset (\mathit{KP}\hspace{.02em}2)\\
S\hspace{-.02em}2 \subset (\mathit{KP}\hspace{.02em}2, \mathit{KM}, \mathit{KN})
\end{array}$\\[.3cm]
\end{tabular}
\end{center}
In the pseudoserendipitous case, a given problem $P1$ in the knowledge
domain $\mathit{KP}1$ becomes solveable (whence, $S\hspace{-.02em}1$) by the addition of
additional knowledge, supplied by $M$.  In the serendipitous case, the
initial set up is similar, but the result is not a solution to the
original problem: rather, it is a new problem, $P\hspace{.02em}2$, together with its
solution.
\vspace{.1cm}\newline\emph{The bridge is transformational.}  Although the
  notation above makes the distinction between the two cases clear, it
  somewhat disguises the principle that is common to
  both.  Even in pseudoserendipity, there's more going on than just new
  information coming online which happens to make a problem solveable.
  Otherwise any online problem-solving system could be seen as
  pseudoserendipitous, which is inconsistent with that term's usage.  When putting together a model
  aeroplane, this is done piece by piece, and even the order in which
  the pieces are put into place is more or less predictable.  It would
  not be said that either the last piece added, nor any of the other
  pieces that were added along the way, was the result of
  pseudoserendipitous creativity.  By contrast, there would have been
  ample opportunity for pseudoserendipity to arise in the historical development of
  powered flight:  \citet[p.~292]{spenser2008airplane} contends that
  ``none of [the progress in aviation] would have happened if human
  interaction hadn't evolved just as dramatically,'' which suggests
  that the process could not have been planned in advance.  To
  consider another example, assembling a jigsaw puzzle is not
  an entirely predictable process: it involves chance at the outset, but
  nevertheless, the overall structure of the solution process is well
  understood.  Even if a previously missing piece was suddenly
  discovered, which made the puzzle solveable, this would not be a
  bridge, because the problem itself is unchanged.  In short, both pseudoserendipitous and serendipitous
  creativity involve ``the transformation of some (one or more)
  dimension of the space so that new structures can be generated which
  could not have arisen before'' \cite[p.~348]{boden1998creativity}.
\vspace{.1cm}\newline\emph{A good problem can be identified by working at a meta-level.}
  The bridge might be thought of as a
  meta-problem, in other words, a fitness function or
  ``aesthetic'' \citep{pease2011computational}, through which an entire
  class of problems may be surveyed, and the most suitable one
  selected (in pseudoserendipity) or induced (in true serendipity).

                                         
\paragraph{\textbf{\upshape Heuristics}} 
~                                         
\newline\emph{Similarity, analogy, and metaphor can be used to retrieve known problems.}
  Instances of  pseudoserendipity concern problems that are known to the system.
  These may be retrieved in a non-transformational way, e.g., via a search process that
  uses analogies between the recently-generated explanation and a catalogue of existing problems.
  \citet{sowa2003analogical} describe three kinds of analogies that
  apply to graphical knowledge structures: matching types with a
  common supertype, matching isomorphic subgraphs, and identifying
  transformations that can change the subgraphs of one graph into
  another.  They give as an example an analogy between a cat and a
  car, found using WordNet data.  In one real-world example, designers
  at Speedo developed a new material to make swimmers faster by
  incorporating a tiny tooth-like network similar to the denticles
  found in the surface of a shark's skin \citep{ingledew2016how}.  The
  concept of ``metaphor'' emphasises the role of a
  representational system in expressing an analogy.
  \citet{xiao2016meta4meaning} describe one way in which the relevant
  background that is needed to interpret (or create) metaphors might
  be acquired.  Structure-based retrieval of source domains may give a
  significant boost to the creativity of the analogies that can be
  constructed \citep{Donoghue2002}.
\vspace{.1cm}\newline\emph{Concept blending may, but does not necessarily, help identify new problems.}
  The bridge might be established by \emph{concept
  blending}, otherwise known as \emph{conceptual integration}
  \citep{fauconnier2008way,fauconnier1998conceptual}.  This approach
  from cognitive science has recently received increased
  attention in computer science
  \citep{confalonieri2018concepts,besold2015analogy,EPPE2018105}.  The
  method forms new combinations of existing concepts---however,
  \citet{fauconnier1998conceptual} advise that ``the most suitable
  analog for conceptual integration is not chemical composition but
  biological evolution.''  Even so, blending can also be
  contrasted with simple models of genetic crossover, where the only
  commonalities that are guaranteed to be preserved are those at at
  the level of individual matching alleles.  In blending, commonalities are
  potentially more abstract.  Finding analogies can be seen as the
  first step in the process of concept blending: for example, given
  the analogy identified by Sowa and Majumdar, multiple different
  cat-car hybrids could be devised, some suitable for nightmares, some
  for children's toys.  Like biological evolution, the blending
  process can involve the outside world in the specification and
  evaluation of blends, and it can do this in ways that combinatorial
  search does not.  \citet{EPPE2018105} have implemented several
  standard-use heuristics that can be used to give basic assessments
  to various blends, but in general blends are evaluated contextually.
  Thagard and Stewart evaluate blends using an abstract simulated
  model of ``cognitive appraisal and physiological perception'' which
  stands for an overall emotional reaction
  \cite[p.~11]{thagard2011aha}. The emotions themselves represent
  circumstances which might be in some sense novel, however they might just as
  well represent a known problem.  Thagard and Stewart
  focus on ``problem solving'' rather than
  problem specification: for them, the ``aha moment'' occurs when there is a
  good match between the newly-generated combination and the
  background emotions.
  Returning to the 3M
  example, sticky notes appeared as a particularly satisfactory blend
  between re-stickable bookmarks and the known problem
  of affixing notes to documents.  The existence
  of the bookmark prototype allowed a new problem to be
  specified: how to attach a note in a way that would not damage the
  document, and would not require a separate fastener.
  This problem likely would never have been considered if
  the only solutions to hand were the previously existing conventional
  technologies of staples, paperclips, and standard-formula glue.  It was an eureka moment for Arthur Fry because he
  had in mind the problem of coming up with a new product: but the
  product itself appeared hand-in-hand with a new problem.  The
  invention of Velcro\textsuperscript{\texttrademark} can similarly be
  reconstructed as a blend, in which the biological problem of
  seed propagation, and its solution of tiny hooks, is blended with
  the domain of fashion to bridge to a new problem: could clothes be
  conveniently fastened using a hook-and-loop mechanism?  We note
  that de Mestral had to expend considerable further effort before he
  was able to answer this question in the affirmative.  This example
  serves to illustrate that a full solution does not always emerge at the same time
  as the problem.
\vspace{.1cm}\newline\emph{Working across domains can give rise to intriguing ideas.} Text mining has
  been used to generate hypotheses by first identifying \emph{bridging
    terms} between different bodies of literature \citep{swanson1997interactive,weeber2001using,jursic2012,jurvsivc2012cross}.
   These methods may be employed in
 \emph{closed discovery} models where
  the ``two domains of interest \ldots\ are identified by the expert prior to starting
the knowledge discovery process'' or
in  \emph{open discovery}
   models where the process works
   ``from a given starting domain towards a yet unknown second domain''
  \citep{jurvsivc2012cross}.
  These correspond, more or less, to the cases of pseudoserendipity and serendipity.
\vspace{.1cm}\newline\emph{Experiments can give surprising insights.}  Experiments have
  been designed using both classic expert system methods
  \citep{Lorenzen1992} as well as modern reinforcement learning
  techniques \citep{melnikov2018active}.  However, it is not clear if
  any software systems are yet looking for bridges between
  experiments, which would allow them to make use of the fact that
  interesting things can be learned when a method is applied ``in just
  a slightly different way'' \cite[p.~28]{austin1978chase}, and
  specialisations of this observation, such as ``the unexpected yield from a control
  experiment may be more fruitful than that from the main experiment''
  (p.~32).

\begin{defn}
\label{def:evaluation}
\hypertarget{def:evaluation}{}\textbf{Evaluation:}
The process results in a product, artefact, process, theory, use for a
material substance, support of a known hypothesis, a solution to a known
problem, a new hypothesis or problem, or some other outcome.
This result is evaluated positively by the system or some external party.
\end{defn}


\paragraph{\textbf{\upshape Foundations}}
~                                         
\newline\emph{Affection is based on reflection.}
\citet{campbell2005serendipity} highlights the idea of ``rational
exploitation'' and the ``discovery of something useful or
beneficial'' as key aspects of serendipity.
But some processing may be required
to get to that point. Here we may refer to the Bergsonian
distinction
between ``perceptions'' and ``affections''
\cite[p.~23]{deleuze1991bergsonism}.
Affection is the ``feeling in the instant'', which is {``}`alloyed'
to other subjectivities [\ldots] as we understand what we feel and
act upon it'' \cite[p.~141]{sutton2008deleuze}.
In particular,
\citet[p.~17]{bergson1991matter} considers affections to be directly
linked to the self-knowledge a being has of its body.  A system's
evaluation of the new state of affairs brought about by the processing
stages outlined in Definitions \ref{def:perception}--\ref{def:bridge}
might be described as ``affective''
when a new system configuration is brought about that is then assessed
in some reflexive way.  Raw somesthetic sense---e.g., an architecture inspired by
the instrumentation of robotic joints with hardwired position
sensors---might be alloyed with ``reflective thinking'' \citep{singh2005architecture}
that considers global aspects of the configuration and course of action
that led to this point.
                                         
\paragraph{\textbf{\upshape Heuristics}} 
~                                         
\newline\emph{Model a sense of taste.} The system's taste is explicitly
  modelled in the case of the artworks evolved by the {\sf Curious
    Design Agents} described by \citet{Saunders2007}.
\vspace{.1cm}\newline\emph{Allow the system to use the world.} As an alternative route to
  working with affect, a system might outsource emotional processing
  to a human user, ``recognise'' the user's affective expression
  \cite[p.~15]{picard1995affective}, and use that as the basis of an
  evaluation.
\vspace{.1cm}\newline\emph{Allow the system to shape its own goals.} 
Whether or not the user
  is given a role in the evaluation process, systems may be designed
  to shape their own goals
  \citep{kaplan2007intrinsically,singh2010intrinsically}.
\subsection{Summary}
We have proposed a phased model of serendipity consisting of several
cognitive components.  We began the section with a schematic diagram
for a computational system that integrates all of these components
(Figure \ref{fig:model}).
We then defined each component with reference to theoretical
literature and existing software implementations, and, where they could
add further clarity, illustrative historical examples.  Table
\ref{tab:model-summary-table} summarises the model that results from
this analysis, highlighting examples of earlier work that support our
definitions and that show the feasibility of the overall proposal.


\begin{landscape}
\tikzset{
    boxComponent/.style={
    rectangle,
    draw=black, very thick,
    minimum height=2em,
    minimum width=.2\linewidth,
    inner sep=2pt,
    text centered,
    },
    boxText/.style={
    rectangle,
    draw=black,
    minimum height=14em,
    minimum width=.2\linewidth,
    inner sep=0em,
    outer sep=0em,
    },
}

\thispagestyle{empty}

\begin{table}
\captionsetup{width=\linewidth}
\vspace{1cm}\hspace{0cm}
\resizebox{1.65\textheight}{!}{
\begin{tikzpicture}[->,>=stealth',node distance=4.2cm]

 \node[boxComponent] (Perception) {\hyperlink{def:perception}{\textbf{Perception}}};
 \node[boxComponent, right of = Perception] (Attention) {\hyperlink{def:attention}{\textbf{Attention}}};
 \node[boxComponent, right of = Attention] (Interest) {\hyperlink{def:focus-shift}{\textbf{Focus Shift}}};
 \node[boxComponent, right of = Interest] (Explanation) {\hyperlink{def:explanation}{\textbf{Explanation}}};
 \node[boxComponent, right of = Explanation] (Bridge) {\hyperlink{def:bridge}{\textbf{Bridge}}};
 \node[boxComponent, right of = Bridge] (Valuation) {\hyperlink{def:evaluation}{\textbf{Valuation}}};


 \coordinate[above= 9mm of Perception  ] (PerceptionPrime) ;
 \coordinate[above= 9mm of Attention    ] (AttentionPrime) ;
 \coordinate[above= 9mm of Interest     ] (InterestPrime) ;
 \coordinate[above= 9mm of Explanation  ] (ExplanationPrime) ;
 \coordinate[above= 9mm of Bridge       ] (BridgePrime) ;
 \coordinate[above= 9mm of Valuation    ] (ValuationPrime) ;

\begin{scope}[thick,decoration={
    markings,
    mark=at position 0.52 with {\arrow{latex}}}
    ]
\draw[postaction={decorate},-] (AttentionPrime) -- (PerceptionPrime);                   
\draw[postaction={decorate},-] (InterestPrime) -- (AttentionPrime);                     
\draw[postaction={decorate},-] (ExplanationPrime) -- (InterestPrime);                   
\draw[postaction={decorate},-] (BridgePrime) -- (ExplanationPrime);                     
\draw[postaction={decorate},-] (ValuationPrime) -- (BridgePrime);                      
\end{scope}
\begin{scope}[thick,decoration={
    markings,
    mark=at position 0.65 with {\arrow{latex}}}
    ]
 \path
 (PerceptionPrime) edge[postaction={decorate},-] (Perception)                                   
 (Attention) edge[postaction={decorate},-,transform canvas={xshift=-9mm}] (AttentionPrime)     
 (AttentionPrime) edge[postaction={decorate},-,transform canvas={xshift=9mm}] (Attention)      
 (Interest) edge[postaction={decorate},-,transform canvas={xshift=-9mm}] (InterestPrime)       
 (InterestPrime) edge[postaction={decorate},-,transform canvas={xshift=9mm}] (Interest)        
 (Explanation) edge[postaction={decorate},-,transform canvas={xshift=-9mm}] (ExplanationPrime) 
 (ExplanationPrime) edge[postaction={decorate},-,transform canvas={xshift=9mm}] (Explanation)  
 (Bridge) edge[postaction={decorate},-,transform canvas={xshift=-9mm}] (BridgePrime)           
 (BridgePrime) edge[postaction={decorate},-,transform canvas={xshift=9mm}] (Bridge)            
 (Valuation) edge[postaction={decorate},-] (ValuationPrime);                                     
\end{scope}


 \node[boxText, below of = Perception,yshift=2em] (PerceptionDef)
 {\small
 \begin{minipage}[t][14em]{10em}
 \begin{flushleft}
  \textbf{Interface to world}\\[.2cm]
  \textbullet\ \citeauthor{russel2003artificial} - different kinds of environments\\
  \textbullet\ \citeauthor{hume1904enquiry}/\citeauthor{peirce1931necessity} - chance is negative/fundamental\\
  \textbullet\ \citeauthor{hoffman2015interface} - adaptivity of not seeing reality as it is\\
  \textbullet\ \citeauthor{friston2009free} - we sense what we can predict
 \end{flushleft}
 \end{minipage}
 };

 \node[boxText, below of = Attention,yshift=2em] (AttentionDef){\small
 \begin{minipage}[t][14em]{10em}
  \begin{flushleft}
  \textbf{Directed processing power}\\[.2cm]
  \textbullet\ \citeauthor{clark2013whatever} - prediction error\\
  \textbullet\ \citeauthor{singh2005architecture} - layered architecture\\
  \textbullet\ \citeauthor{bateson-logical-categories} - changing behaviour\\
  \textbullet\ \citeauthor{rowley2007wisdom} - meaning making
  \end{flushleft}
 \end{minipage}
 };

 \node[boxText, below of = Interest,yshift=2em] (InterestDef) {\small
 \begin{minipage}[t][14em]{10em}
\begin{flushleft}
  \textbf{Evaluation of data via existing objective functions}\\[.2cm]
  \textbullet\ Wundt curve\\
  \textbullet\ \citeauthor{berlyne1954theory} - epistemic and perceptual curiosity\\
  \textbullet\ \citeauthor{logan1994modelling} - belief revision in information seeking\\
  \textbullet\ \citeauthor{patalano1993predictive} - predictive encodings
\end{flushleft}
 \end{minipage}
 };

 \node[boxText, below of = Explanation,yshift=2em] (ExplanationDef){\small
 \begin{minipage}[t][14em]{10em}
\begin{flushleft}
  \textbf{A predictive model}\\[.2cm]
  \textbullet\ \citeauthor{lawson1998metaphysics} - principles and causes\\
  \textbullet\ \citeauthor{pease2011computational} - framing\\
  \textbullet\ \citeauthor{bateson-logical-categories} - change of pattern\\
\end{flushleft}
 \end{minipage}
 };

 \node[boxText, below of = Bridge,yshift=2em] (BridgeDef) {\small
 \begin{minipage}[t][14em]{10em}
\begin{flushleft}
  \textbf{Identifying or positing a problem (via a new objective function)}\\[.2cm]
  \textbullet\ \citeauthor{bergson1946creative} - creativity of problem statement\\
  \textbullet\ \citeauthor{thagard2011aha} - ``aha moment''\\
  \textbullet\ \citeauthor{boden1998creativity} - transform the space\\
  \textbullet\ \citeauthor{pease2011computational} - new aesthetic
\end{flushleft}
 \end{minipage}
 };

 \node[boxText, below of = Valuation,yshift=2em] (ValuationDef) {\small
 \begin{minipage}[t][14em]{10em}
\begin{flushleft}
  \textbf{Evaluation of solution via existing objective function}\\[.2cm]
  \textbullet\ \citeauthor{bergson1991matter} - affection\\
  \textbullet\ \citeauthor{campbell2005serendipity} - rational exploitation\\
\end{flushleft}
 \end{minipage}
 };

 \node[boxText, below of = PerceptionDef, yshift = -4em] ()
 {\small
 \begin{minipage}[t][14em]{10em}
\begin{flushleft}
  \textbf{HCI, automated feature finding, emergence of grid cells} \\[.2cm]
  \textbullet\ \citeauthor{turk2000perceptive}\\
  \textbullet\ \citeauthor{jacob2015viewpoints}\\
  \textbullet\ \citeauthor{stopher2017technology}\\
  \textbullet\ \citeauthor{inceptionism}\\
  \textbullet\ \citeauthor{Banino2018}
\end{flushleft}
 \end{minipage}
 };

 \node[boxText, below of = AttentionDef, yshift = -4em] ()
 {\small
 \begin{minipage}[t][14em]{10em}
\begin{flushleft}
  \textbf{Visual attention, competition for resources, temporal bonus, soft attention}\\[.2cm]
  \textbullet\ \citeauthor{sun2003object}\\
  \textbullet\ \citeauthor{tsotsos1995modeling}\\
  \textbullet\ \citeauthor{baars1997theatre}\\
  \textbullet\ \citeauthor{lesser1977retrospective}\\
  \textbullet\ \citeauthor{vemula2017social}
\end{flushleft}
 \end{minipage}
 };

 \node[boxText, below of = InterestDef, yshift = -4em] ()
 {\small
 \begin{minipage}[t][14em]{10em}
\begin{flushleft}
  \textbf{Autonomous creative behaviour, aesthetics classifier, compression, information gain}\\[.2cm]
  \textbullet\ \citeauthor{Saunders2007}\\
  \textbullet\ \citeauthor{dhar2011high}\\
  \textbullet\ \citeauthor{schmidhuber2009art}\\
  \textbullet\ \citeauthor{javaheri2016analysis}
\end{flushleft}
 \end{minipage}
 };

 \node[boxText, below of = ExplanationDef, yshift = -4em] ()
 {\small
 \begin{minipage}[t][14em]{10em}
\begin{flushleft}
  \textbf{Explanation-based learning, epistemic modelling, critics, dialogue, integration of causal models}\\[.2cm]
  \textbullet\ \citeauthor{ellman1989explanation}\\
  \textbullet\ \citeauthor{delamaza1994generate}\\
  \textbullet\ \citeauthor{sussman1973computational}\\
  \textbullet\ \citeauthor{singh2005alternate}\\
  \textbullet\ \citeauthor{moore1995participating}\\
  \textbullet\ \citeauthor{GeiHofSch16}
\end{flushleft}
 \end{minipage}
 };

 \node[boxText, below of = BridgeDef, yshift = -4em] ()
 {\small
 \begin{minipage}[t][14em]{10em}
\begin{flushleft}
  \textbf{Analogy, metaphor, concept blending, bridging terms}\\[.2cm]
  \textbullet\ \citeauthor{sowa2003analogical}\\
  \textbullet\ \citeauthor{xiao2016meta4meaning}\\
  \textbullet\ \citeauthor{confalonieri2018concepts}\\
  \textbullet\ \citeauthor{EPPE2018105}\\
  \textbullet\ \citeauthor{swanson1997interactive}\\
  \textbullet\ \citeauthor{jursic2012}
\end{flushleft}
 \end{minipage}
 };

 \node[boxText, below of = ValuationDef, yshift = -4em] ()
 {\small
 \begin{minipage}[t][14em]{10em}
\begin{flushleft}
  \textbf{Modelling taste, affect, intrinsic motivation}\\[.2cm]
  \textbullet\ \citeauthor{Saunders2007}\\
  \textbullet\ \citeauthor{picard1995affective}\\
  \textbullet\ \citeauthor{kaplan2007intrinsically}\\
  \textbullet\ \citeauthor{singh2010intrinsically}
\end{flushleft}
 \end{minipage}
 };

 \node [rotate=90, left of = Perception, xshift=2.5em,yshift=7em] {\textbf{Definitions}};

  \node [rotate=90, left of = Perception, xshift=-14em,yshift=7em] {\textbf{Implementations}};

\begin{scope}[thick]
 \path
 (Perception) edge (Attention)
 (Attention) edge (Interest)
 (Interest) edge (Explanation)
 (Explanation) edge (Bridge)
 (Bridge) edge (Valuation)
;
\end{scope}
\end{tikzpicture}}

\begin{tabular}{p{.1\textwidth}p{.\textwidth}}
\caption{Our model for systems with serendipity potential. The flowchart at top provides a visual key, showing that previous phases can be returned to at any point. The body of the table summarises Definitions \ref{def:perception}--\ref{def:evaluation}, with references to previous models and existing implementations per component.\label{tab:model-summary-table}}
\end{tabular}
\end{table}
\end{landscape}

\section{Testing the effectiveness of the model: Can it discriminate between systems that have serendipity potential and those that do not?}\label{sec:system-analysis}

Here we test the effectiveness of our model at discriminating between
systems that have previously been described as (in some sense)
serendipitous, and one example of a system that seems to be decidedly
non-serendipitous.  If the model can achieve this, that should increase our
confidence that the model outlines an implementable
characterisation of a system's serendipity potential. 

The systems we examine are:
\begin{itemize}
\item[] Mueller's {\sf DAYDREAMER} -- serendipity was a key concern its design (\cite{mueller1990}, \textsection 5.3): will our model affirm that it has serendipity potential?
\item[] A pocket calculator -- such a simple system seems intuitively unlikely to exhibit features of serendipity: will our model reproduce this result?
\item[] Pask's {\sf Colloquy of Mobiles} -- this was an interactive system that was designed with some notion of serendipity in mind \citep{pask1971comment}: what can our model say about the relationship between serendipity in the system and serendipity as a service in this case?
\item[] Ramezani's {\sf GH} -- this is a contemporary discovery system that did not explicitly consider serendipity in its design \citep{ramezani2014artificial}: its serendipity potential was assessed by \citet{pease2013discussion} using their evaluation framework, and it appears to be something of an edge case.  Can our more refined model yield a decisive ruling?
\end{itemize}
We code each facet as YES,
NO, or SOME, depending on the presence, absence, or partial presence
of indicators matching the definitions and heuristics given in Section
\ref{sec:our-model}.


%

\subsection{{\sf DAYDREAMER}}
The {\sf DAYDREAMER} system \cite{mueller1990} is intended to provide a computational model of daydreaming.  An agent is guided to use its `imagination' to develop ideas and construct short narratives.  The principle behind {\sf DAYDREAMER} is that a planning agent can operate in a `relaxed' manner to explore possibilities in unusual ways, where the relaxation state is achieved by removing or reducing constraints on the search process that guides the exploration.  {\sf DAYDREAMER}'s exploration is driven by loosely constrained planning mechanisms which are given a pre-determined goal.
The generated plan then becomes the basis of a narrative.
Mueller identifies a distinction between {\sf DAYDREAMER} and other comparable systems:
\begin{quote}
\emph{``There are certain needless limitations of most present-day
  artificial intelligence programs which make creativity difficult or
  impossible: They are unable to consider bizarre possibilities and
  they are unable to exploit accidents.''} \cite[p. 14]{mueller1990}
\end{quote}
In other words, the {\sf DAYDREAMER} system was designed to capitalise
on the unusual or accidental {\em non-obvious} options available to
it, which gives intuitive support for Mueller's case that it can act
serendipitously.  We apply our model to check whether this claim is
justified: details are given in Table \ref{ex:daydreamer}.

\let\oldparagraph\paragraph
\makeatletter
\renewcommand{\paragraph}{%
  \@startsection{paragraph}{4}%
  {\z@}{1ex \@plus 1ex \@minus .2ex}{-1em}%
  {\normalfont\normalsize\itshape}%
}
\makeatother

\begin{table}
\begin{mdframed}[innertopmargin=-5pt]
\begin{tabular}{p{.96\textwidth}}
\paragraph{Perception:}
\textbf{SOME.} {\sf DAYDREAMER} has access to the outside world in that it
can be given information about events, physical objects and goals as
input. However, it lacks perception of events beyond such input,
and cannot steer its perception, or significantly structure the input.
\paragraph{Attention:}
\textbf{YES.} {\sf DAYDREAMER} is able to direct processing
power to pay attention to different aspects of perception.  It is able
to interpret input in the context of domain knowledge, and also in
light of previous daydreams.
\paragraph{Focus shift:}
\textbf{YES.} Information is processed and evaluated according to an
emotional component and personality traits implemented within
{\sf DAYDREAMER}, which determine what {\sf DAYDREAMER} does and
does not take note of.  Focus shifts are targeted towards achieving
a particular goal.  The system has an explicit notion of contexts and shifts between them:
``planning rules give rise to alternative states of a hypothetical world''
\cite[p.~35]{mueller1990}.
\paragraph{Explanation:}
\textbf{YES.}  Drawing on previous experience and domain knowledge
       {\sf DAYDREAMER}, regularly executes a `predictor' function to
       measure whether new conceptual steps are likely to bring it
       closer to its goal.
\paragraph{Bridge:}
\textbf{YES.}  {\sf DAYDREAMER} can employ analogical reasoning to see
if aspects of the plan it is working on could be adapted to achieve
some other existing goal.  It can also retrieve and reuse the plans
it has previously created.
\paragraph{Valuation:}
\textbf{SOME.} Valuation is performed by {\sf DAYDREAMER} by assessing whether
the goal it is trying to achieve has been realised.  There is no valuation of the interestingness
or variability of the daydreams produced over time. 
The system has limited ability to select topics to daydream about next.
\end{tabular}
\end{mdframed}
\vspace{-.2cm}
\caption{Applying our model to evaluate the serendipity potential of the {\sf DAYDREAMER} system\label{ex:daydreamer}}
\end{table}

\begin{table}
\begin{mdframed}[innertopmargin=-5pt]
\begin{tabular}{p{.96\textwidth}}
\paragraph{Perception}
\textbf{SOME.} A calculator has the ability to perceive any input that
is given to it by the user.  However, it has no other mechanisms for
perception of the outside world.
\paragraph{Attention}
\textbf{NO.} A calculator pays attention equally to every input,
with no ability to discern one element over another above basic
sequential processing involved in calculations.  In principle,
a limited exception might be provided by a `memory', `M', or `mem' key, which
stores particular numbers upon a user request (i.e., by pressing the
key). It could be argued that the calculator is paying
particular attention to the value(s) stored in memory: however
since this is entirely directed by the user, not the calculator,
we do not consider this to match our definition of attention.
\paragraph{Focus shift through interest}
\textbf{NO.} The calculator evaluates data via
functions, however these are not ``objective functions,''
since the calculator has no goals.
Even when encountering an error, a calculator does not effect a focus shift.
\paragraph{Explanation}
\textbf{NO.} A scientific calculator might record a log of its work,
but would not explain the process or any aspect thereof.
\paragraph{Bridge}
\textbf{NO.} Calculators solve mathematical problems, one at a time;
they cannot extrapolate to solve other problems which have not been
posed to them. 
\paragraph{Valuation}
\textbf{NO.} A calculator has no concept of evaluating the correctness
or fitness of solutions it generates; it merely provides the one solution
that it has been programmed to generate.  It also cannot evaluate its
processes or strategies.
\end{tabular}
\end{mdframed}
\vspace{-.2cm}
\caption{Applying our model to evaluate the serendipity potential of a pocket calculator\label{ex:calculator}}
\end{table}

Although there are some dimensions where {\sf DAYDREAMER} could be
strengthened in order to have more serendipity potential, notably in
its perception abilities and its valuation of what it does, the system
is overall a good demonstration of our model.  Symmetrically, the
model shows good evidence to support Mueller's assertion that the
system does have serendipity potential.  Furthermore, the system
appears to be able to manifest both pseudoserendipity and serendipity
proper, as illustrated by these two examples:
\begin{enumerate}[label=(\roman*)]
\item ``\emph{{\sf DAYDREAMER} receives an alumni directory from the college
    she attended which happens to contain the number of Carol Burnett.
    {\sf DAYDREAMER} had previously been daydreaming about contacting
    Harrison Ford in order to ask him out again.  \ldots\
    {\sf DAYDREAMER} realizes that the alumni directory is applicable
    to the problem of finding out the unlisted telephone number of
    Harrison Ford.  {\sf DAYDREAMER} could possibly find out
    Harrison's telephone number by obtaining a copy of the alumni
    directory from the college Harrison Ford attended, if any.}''
  \cite[p.~125]{mueller1990}.
\item ``[S]\emph{uppose {\sf DAYDREAMER} is again concerned with how
  to meet Harrison Ford when it happens to have a car accident.  As
  {\sf DAYDREAMER} is exchanging telephone numbers and other
  information with the person, it notices that one way of meeting
  Harrison Ford is to force an accident with him.  The next time the
  program has the goal of meeting someone, the plan of forcing an
  accident with that person will immediately be retrieved.  This
  solution is one which would have been difficult to generate out of
  thin air.}  \cite[p.~126]{mueller1990}.
\end{enumerate}


\subsection{Calculator} \label{sec:calculator}

Having applied our model as above a system that could reasonably be
described as serendipitous, we now seek to check whether the model is
effective in ruling out non-serendipitous systems.  Or might it yield
false positives?  We consider the example of a pocket calculator
(Table \ref{ex:calculator}).\footnote{It seems likely that a
  calculator could be successfully used as part of a system delivering
  serendipity as a service, for instance as a source of random
  numbers, but we focus here on checking for serendipity in the
  system.}

Since it has an interface to the outside world, the calculator
matches our definition of perception, 
however, it is a poor match for the remaining features of our model.
Thus, the model is effective in showing no serendipity potential in a
calculator, as we had hoped.

\begin{table}
\begin{mdframed}
\begin{tabular}{@{\hspace{-0.01\textwidth}}p{.5\textwidth}@{\hspace{.02\textwidth}}p{.5\textwidth}}
\begin{minipage}{.48\textwidth}
\vspace{-3.95\baselineskip}
\textbf{[System]}
\paragraph{Perception:} \textbf{SOME.}
The mobiles were given light and sound sensors, which are linked to their drives.
The mobiles' behaviour is controlled by light and sound behaviour in their
environment, which can originate from other mobiles or from other sources.
\paragraph{Attention:} \textbf{SOME.}
The mobiles have ``gender roles'' which cause them to turn to one another
looking for certain behaviours to satisfy their drives.  They are, however,
given only limited attention spans.
\paragraph{Focus shift:} \textbf{SOME.} Once attention
has been captured, a mobile will change its behaviour until its drive
is satisfied or interrupted.
\paragraph{Explanation:}
\textbf{SOME.}  The female mobiles ``were adaptive in the sense that
they could learn to identify individual males and remember their
peculiarities''  \citep{pickering2007ontological}.
\paragraph{Bridge:} \textbf{NO.} The mobiles did not have
the ability to identify any problems other than the satisfaction of
their drives, nor could they strategise about how to satisfy those
drives beyond the simple form of learning mentioned above.
\paragraph{Valuation:} \textbf{NO.} While the mobiles continuously performed
local optimisations, there was no ``result'' that could be valued (nor were
they given the ability to form valuations).
\end{minipage} &
\begin{minipage}{.48\textwidth}
\textbf{[Audience]}
\paragraph{Perception:} \textbf{YES.}
Audience members were able to perceive the installation as a whole,
and also interact with it using light and sound (and perceive the
effects of their own interactions). 
\paragraph{Attention:} \textbf{YES.}
The museum-going public also has limited attention spans.
\paragraph{Focus shift:} \textbf{YES.}
Pask notes in an appendix 
that audience members interacted interestedly with the system \cite[p.~98]{pask1971comment}.
\paragraph{Explanation:} \textbf{YES.}
Audience members were able to generate theories about how
their ``actions lead to impacts on the environment that lead to
sensing and further motivation of actions'' by the mobiles
\citep{haque2007architectural}.
\paragraph{Bridge:} \textbf{SOME.} 
At least some commentators were able to abstract from the exhibit to further philosophical thinking
about ``what sorts of things there are in the world, and how they
relate to one another'' \citep{pickering2007ontological}.
``Conversational machines'' were not part of everyday life in 1968,
and the system can still provoke debate \citep{pangaro2018serendipity}.
\paragraph{Valuation:} \textbf{SOME.}
\citet[p.~5]{gemeinboeck2015performance} remark: ``The work introduced machinic attributes that even
today still sound very advanced to museum audiences'' and it ``is
in many ways as much a humorous, social observation of humans and
their nonhuman counterparts as it is a technological achievement.''
\end{minipage}
\end{tabular}
\end{mdframed}
\caption{Applying our model to evaluate the serendipity potential of Pask's {\sf Colloquy of Mobiles}.  The system itself is evaluated in the left column, whereas the audience's experience of the system is evaluated in the right column. \label{ex:mobiles}}
\end{table}

\subsection{{\sf Colloquy of Mobiles}}

Gordon Pask's {\sf Colloquy of Mobiles} was one of the installations
that appeared in the 1968 Cybernetic Serendipity exhibition at the
Institute of Contemporary Arts in London
\citep{reichardt1969cybernetic}.  The exhibition itself proved popular
with the museum-going public at the time, and has been extensively
discussed in subsequent literature
\citep{Edmonds1994,macgregor2002cybernetic,usselmann2003dilemma}.  For
our purposes the interesting question is whether, and how, the concept
of ``serendipity'' relates to one of the more famous artworks that was
exhibited.

In an essay that describes the details of his installation, composed
before the exhibition took place, Pask wrote:

\begin{quote}
``[T]\emph{he mobiles produce a complex auditory and visual effect by
  dint of their interaction.  They cannot, of course, interpret these
  light and sound patterns.  But human beings can and it seems
  reasonable that they will also aim to achieve patterns that they deem
  pleasing by interacting with the system at a higher level of
  discourse.}
\emph{I do not know.  But I believe it may work out that way.}''  \cite[p.~91]{pask1971comment}
\end{quote}

While the system components have been given regulatory goals which are
realised in a
stochastic way, the
system components are not themselves able to make any deeper sense of
their communication or behaviour.  This suggests that we should make a
dual accounting, and examine the potential for serendipity on the
side of the system, and compare it with the potential for serendipity
on the side of the audience  (Table \ref{ex:mobiles}).
According to our analysis, there was no possibility for serendipity on
the system side, but nevertheless there was a possibility for
serendipity in the ``wider'' system that included human actors.


\begin{table}[t]
\begin{mdframed}[innertopmargin=-5pt]
\begin{tabular}{p{.96\textwidth}}
\paragraph{Perception:} \textbf{SOME.}
 New data comes online in a given Dynamic Investigation Problems, and {\sf GH} has a memory of
 previous DIPs.
\paragraph{Attention:} \textbf{SOME.}
 Search/inference operates with a limited scope.
\paragraph{Focus shift:} \textbf{YES.}
{\sf GH} can achieve a focus shift
 ``if a previous case is re-evaluated by the system as relevant to
 the current case'' \cite[p.~67]{pease2013discussion}.
\paragraph{Explanation:} \textbf{YES.}
 The system can produce a proof demonstrating certain conclusions
 (e.g., the culprit in a Cluedo-style mystery or the likely cause
 of a disease).  This is a predictive model and thus an explanation
 in our sense of the word.
\paragraph{Bridge:} \textbf{NO.} 
 The system can build an expanded solution strategy by using previously
 solved problems to flesh out its current challenge, however this does
 not amount to either problem identification or problem creation.
\paragraph{Valuation:} \textbf{SOME.}
 The system can assign
 likelihood to a given solution or diagnosis in an online fashion: its
 confidence in the solution could be understood as the solution's value.
\end{tabular}
\end{mdframed}
\caption{Applying our model to evaluate the serendipity potential of Ramezani's {\sf GH} system\label{ex:gh}}
\end{table}

\subsection{The {\sf GH} System}

\citet{pease2013discussion} assessed the {\sf GH} system developed
by \citet{ramezani2014artificial}.  It met
almost all the criteria for serendipitous behaviour advanced in
their paper.  In brief, {\sf GH} solves Dynamic Investigation Problems
(DIPs), similar to the tabletop mysteries that unfold in the board
game Cluedo (Clue, in North America).  However, {\sf GH} fails to meet
two environmental criteria advanced by Pease et al: ``it only solves
one \emph{task} at a time, and there are not currently \emph{multiple
  influences}'' (p.~67, emphasis in original).  As we see in Table
\ref{ex:gh}, the system may be understood to meet many of our current
criteria in at least a partial sense, but it fails to achieve a bridge
as this concept is understood in Definition \ref{def:bridge}.

Path dependence of a solution---in which a system happens to have the
relevant preparations to solve a given problem---is not the same as
serendipity.
\citet{Campos2002} allow the transformation of a known but unsolvable
problem into a solvable one, through the use of data acquired in an
online fashion, to be termed ``pseudoserendipity.''  With Definition
\ref{def:bridge}, we aim to be more stringent, and foreground the
nontrivial nature of the transformation.
In our assessment, while {\sf GH} attempts to solve dynamic problems,
and makes use of a memory of related problems to help solve them, it
only exhibits path dependence, not bridging, since it does not use
online data to transform its problems, or its approach to solving
them.

In principle, the system could be restructured to have an ongoing set
of ``cases'' that it revisits periodically, and whereby online
learning sparked in one case may (pseudoserendipitously) be bridged to
solutions in other cases.  This redesign would be representative of
the \emph{multiple tasks} criterion from \citet{pease2013discussion},
who discussed a similar learning architecture for a different system.


\let\paragraph\oldparagraph

\subsection{Summary}
As Table \ref{tab:systems-analysis} shows, our model
can effectively discriminate between systems that have little or no
potential to be serendipitous, and computational or interactive
systems that possess serendipity potential.  
\begin{itemize}
\item {\sf DAYDREAMER} meets our criteria for \emph{serendipity in the
  system}, though two are only met weakly.
\item {\sf Colloquy of Mobiles} meets the criteria only when viewed as a
  system for \emph{serendipity as a service.}
\end{itemize}
Our ruling is that {\sf GH} fails to meet the full requirements of the
model.  Future work might address the deficit by exploring how online
learning in the context of Dynamic Investigation Problems could be
applied to as-yet-unencountered problems; or, pseudoserendipitously,
if strategies used to solve new DIPs yielded insights about how to
solve known but previously-insoluble DIPs.

The serendipity potential of {\sf DAYDREAMER} and {\sf Colloquy of
  Mobiles} might be increased in further rounds of prototyping.
The source code for {\sf DAYDREAMER} is
online,\footnote{\url{https://github.com/eriktmueller/daydreamer}} and
\citet{pangaro2018serendipity} are building {\sf Colloquy of Mobiles
  2018} using contemporary technologies, intending to ``open-source
everything found and everything generated, including CAD numerical
models and engineering drawings''---so such progress may indeed be
possible.

\begin{table}[ht]
\begin{center}
\footnotesize

\small{
\begin{tabularx}{\textwidth}{l|p{\widthof{Expla}}|p{\widthof{Expla}}|p{\widthof{Expla}}|p{\widthof{Explaion}}|p{\widthof{Expla}}|p{\widthof{Expla}}|}
\multicolumn{1}{c}{\phantom{Systemxxxxxxx}} &
\multicolumn{1}{c}{Perception} &
\multicolumn{1}{c}{\hspace{-.7em}Attention} & 
\multicolumn{1}{c}{F/Shift} &
\multicolumn{1}{c}{\hspace{-.4em}Explanation} &
\multicolumn{1}{c}{\hspace{-.4em}Bridge} &
\multicolumn{1}{c}{Evaluation} \\
\end{tabularx}}

\begin{tabularx}{\textwidth}{l|p{\widthof{Expla}}|p{\widthof{Expla}}|p{\widthof{Expla}}|p{\widthof{Explaion}}|p{\widthof{Expla}}|p{\widthof{Expla}}|}
\hline 
\multicolumn{1}{l}{}&\multicolumn{1}{c}{}&\multicolumn{1}{c}{}&\multicolumn{1}{c}{}&\multicolumn{1}{c}{}&\multicolumn{1}{c}{}&\multicolumn{1}{c}{}\\
\hhline{~------}
\multicolumn{1}{l}{\scriptsize {\sf DAYDREAMER}} & \multicolumn{1}{|Y|}{\cellcolor{yellow!25}SOME} & \multicolumn{1}{|Y|}{\cellcolor{green!25}Y}    & \multicolumn{1}{|Y|}{\cellcolor{green!25}Y}    & \multicolumn{1}{|Y|}{\cellcolor{green!25}Y}    & \multicolumn{1}{|Y|}{\cellcolor{green!25}Y}    & \multicolumn{1}{|Y|}{\cellcolor{yellow!25}SOME} \\
\hhline{~------}
\multicolumn{1}{l}{\scriptsize Calculator}            & \multicolumn{1}{|Y|}{\cellcolor{yellow!25}SOME} & \multicolumn{1}{|Y|}{\cellcolor{red!25}N} & \multicolumn{1}{|Y|}{\cellcolor{red!25}N}    & \multicolumn{1}{|Y|}{\cellcolor{red!25}N}    & \multicolumn{1}{|Y|}{\cellcolor{red!25}N}    & \multicolumn{1}{|Y|}{\cellcolor{red!25}N}    \\
\hhline{~------}
\multicolumn{1}{l}{\scriptsize {\sf C}.-{\sf M}. (System)}   & \multicolumn{1}{|Y|}{\cellcolor{yellow!25}SOME} & \multicolumn{1}{|Y|}{\cellcolor{yellow!25}SOME} & \multicolumn{1}{|Y|}{\cellcolor{yellow!25}SOME} & \multicolumn{1}{|Y|}{\cellcolor{yellow!25}SOME} & \multicolumn{1}{|Y|}{\cellcolor{red!25}N}    & \multicolumn{1}{|Y|}{\cellcolor{red!25}N}    \\
\hhline{~------}
\multicolumn{1}{l}{\scriptsize {\sf C}.-{\sf M}. (Audience)} & \multicolumn{1}{|Y|}{\cellcolor{green!25}Y}    & \multicolumn{1}{|Y|}{\cellcolor{green!25}Y}    & \multicolumn{1}{|Y|}{\cellcolor{green!25}Y}    & \multicolumn{1}{|Y|}{\cellcolor{green!25}Y}    & \multicolumn{1}{|Y|}{\cellcolor{yellow!25}SOME} & \multicolumn{1}{|Y|}{\cellcolor{yellow!25}SOME} \\
\hhline{~------}
\multicolumn{1}{l}{\scriptsize {\sf GH}}   & \multicolumn{1}{|Y|}{\cellcolor{yellow!25}SOME} & \multicolumn{1}{|Y|}{\cellcolor{yellow!25}SOME} & \multicolumn{1}{|Y|}{\cellcolor{green!25}Y} & \multicolumn{1}{|Y|}{\cellcolor{green!25}Y} & \multicolumn{1}{|Y|}{\cellcolor{red!25}N} & \multicolumn{1}{|Y|}{\cellcolor{yellow!25}SOME}   \\
\hhline{~------}
\end{tabularx}
\normalsize
\end{center}
\caption{Summary of our analysis of the serendipity potential of example systems: {\sf DAYDREAMER} arguably meets our criteria for serendipity in the system;
{\sf Colloquy of Mobiles} ({\sf C}.-{\sf M}.) meets the criteria only when viewed as a system for \emph{serendipity as a service}; a pocket calculator is missing most of the features; {\sf GH} is missing the bridge facet. \label{tab:systems-analysis}}
\end{table}

\section{{\sf HR} and {\sf HRL}: On the trail of serendipity} \label{sec:pursuit}
In this section we give an account of several
  episodes in a historical sequence of development of the related
  discovery systems {\sf HR} and {\sf HRL}, developed by two of us
  (Colton and Pease). We use our framework to discern the serendipity
  potential, if any, for the systems described at various stages of
  development.  Our account illustrates that by combining rich domain
  knowledge and reasoning methods, the dimensions of our framework can
  be brought online in applied domains.  Given the nature of the
  systems discussed, throughout this section the ideal
  ``explanation'' is a mathematical proof, though other forms of explanation are seen to be relevant.  In Table \ref{hr-episodes-table} we will zoom in on the presence or absence of focus shifts in these episodes.

\begin{ep}[{\sf HR} constructs the concept of the central elements in a group]\label{ex:central}
The {\sf HR} system\footnote{Named after mathematicians Hardy (1877 - 1947) and Ramanujan (1887 - 1920).} \citep{colton2002automated} is a machine learning tool which performs automated discovery in a variety of domains. HR starts with objects of interest (such as integers) and initial concepts (such as division, multiplication and addition) and uses production rules to transform either one or two existing concepts into new ones. HR also makes conjectures which empirically hold for the objects of interest supplied, and has a set of interestingness measures which it uses to evaluate its new concepts and conjectures.

One early success was in the domain of abstract algebra, in which {\sf HR} developed a trigger concept to re-discover the concept of \emph{the central elements of a group} (the set of elements in a group that commute with every element in the group) \citep{colton2002automated}. Here the trigger was the concept [$a$,$b$,$c$] : $a*b=c$. Having \emph{perceived} this concept (which it generated), HR gives it further \emph{attention}, by first evaluating it (positively), in the context of its objects of interest, the other concepts in the theory, its conjectures, and so on.  The concept is then recontextualised through the application of HR's {\em compose}, {\em exists} and {\em forall} production rules in the following way:
\begin{align*}
\left.
\begin{array}{c}
{[}a,b,c{]} : a*b=c\\
{[}a,b,c{]} : a*b=c
\end{array}
\right\}\:\:
&\mathbf{compose}\rightarrow [a,b,c] : a*b=c \:\text{\emph{\&\&}}\: b*a=c\\
&\mathbf{exists}\rightarrow [a,b] : \text{exists}\ c\ (a*b=c \:\text{\emph{\&\&}}\: b*a=c)\\
&\mathbf{forall}\rightarrow [a] : \text{all}\ b\ (\text{exists}\ c\ (a*b=c \:\text{\emph{\&\&}}\: b*a=c))
\end{align*}
Thus, by building on the notion of multiplication in a group, HR has (re)disco-vered concept of \emph{the central elements of a group}.  The \emph{evaluation} of this concept is positive, as judged independently both by HR and externally, by virtue of being recognised as a core concept in Group Theory, and appearing in most if not all basic textbooks on the subject. This renders multiplication itself more interesting as a potential source for further concepts.   However, as it happens the concept of multiplication did not become the more interesting simply because it is used to form an interesting concept, so no \emph{focus shift} takes place.  Similarly, the \emph{explanation} and \emph{bridge} criteria are not met in this iteration of the system.
\end{ep}

\begin{ep}[{\sf HR} refutes a boring conjecture in monoid theory]\label{ex:monoid}
Colton subsequently enhanced the system so that whenever it finds a
counterexample to a new conjecture, it tests to see whether the
counterexample also breaks some other previously unsolved open conjecture.
In this case, the system's ``prepared mind'' takes the form of previous
experiences, background knowledge, a store of unsolved problems, as
well as skills and a current focus.  The new counterexample arises
partly due to factors beyond the system's control, in particular, the
built-in structure of the domain.

This version of the system was tested in three test domains:  group theory (associativity, identity and inverse axioms), monoid theory (associativity, identity) and semigroup theory (associativity). When {\sf HR} runs in breadth first mode, i.e., applying all production rules in order without any heuristic search, then during sessions with tens of thousands of production rule steps, there were no instances of open problems which were solved in this way. Amending the search strategy to randomly select one of the available production rules led to one instance of a newly generated counterexample solving a pre-existing conjecture in monoid theory, none at all in group theory and a handful of times in semi-group theory (there were three times when a new counterexample dispatched an open conjecture, and on one occasion, ten open conjectures were dispatched by one counterexample).  However, not only was this a rare occurrence, but the conjectures which were disproved in this way could not be considered interesting: for instance, the monoidal conjecture disproved by a later counterexample was the following:
\begin{align*}
\forall b, c, d &\hspace{.2cm}(((b * c = d \wedge c * b = d \wedge c * d = b \wedge (\exists(e * c = d \wedge e * d = c)))\\
&\hspace{-.2cm}\leftrightarrow(b * c = d \wedge (\exists f(b * c = f)) \wedge (\exists g(g * c = b)) \wedge d * b = c \wedge c * d = b)))
\end{align*}

\noindent This conjecture does {\em not} appear in textbooks on Monoid Theory.

Alongside the attributes of \emph{perception} and \emph{attention} as
described in in Episode \ref{ex:central}, it seems we may now have a
evidence of a focus shift, since open conjectures are
reconsidered in light of a potential counterexample.
However, we must be careful with our analysis.
In this case, a potentially interesting open conjecture
becomes uninteresting once it has been refuted.
That is to say, its evaluation goes down, so the
precondition for a focus shift is not present.  Neither
is there at any stage a reevaluation of the counterexample itself.
So, again, no \emph{focus shift} takes place.

However, the refutation of a conjecture does constitute an \emph{explanation},
since it proves the conjecture's falsity.  The results obtained,
as illustrated by the example given above, were never \emph{bridged} to further problems.
The \emph{evaluation} of the refuted conjecture, as judged both internally by HR and externally, is low.
\end{ep}

\begin{ep}[{\sf HRL} undiscovers the platypus]\label{ex:platypus}
{\sf HRL} was an adaptation of {\sf HR}, developed by \citet{pease07}
and based on a theory of argumentation that acknowledges the role of
conflict and ambiguity in mathematical discovery.  The theory, based
on the work of \citet{lakatos}, can also be used to describe (some)
real-world discoveries in mathematics.  {\sf HRL} is a distributed
system, comprised of ``student'' and ``teacher'' agents, each running
a copy of Colton's {\sf HR}.  The agents all have a similar
architecture, but different input knowledge, measures of
interestingness, and different ways of producing concepts.  The
overall system is organised into work phases and discussion phases, in
which conjectures, concepts, and counterexamples are communicated.
Students react to counterexamples using Lakatos's methods.  One such
discussion, developed around a simple theory of animals, progressed as
follows:
\begin{itemize}
\item[\emph{A}:] ``There does not exist an animal which produces milk and lays eggs.''
\item[\emph{B}:] ``The platypus does.''
\item[\emph{A}:] {[}Checks new object against current theory. Finds it breaks 11\% of its conjectures.{]}\newline ``The platypus is not an animal.''
\item[\emph{B}:] {[}Finds that the platypus breaks 31\% of its own conjectures.{]}\newline ``Okay - I'll accept that.''
\end{itemize}
\end{ep}

We will discuss this example together with the following:

\begin{ep}[{\sf HRL} formulates Goldbach's Conjecture]\label{ex:goldbach}
The same system could also do theory formation in basic number theory.
Here is another dialogue:
\begin{itemize}
\item[\emph{A}:] {[}Knows: numbers 10-20, integer, div, mult{]}\newline ``All even numbers are the sum of two primes.''
\item[\emph{B}:] {[}Knows: numbers 0-10, integer, div, mult{]}\newline ``2 is not the sum of two primes.''
\item[\emph{A}:] {[}Checks new object against current theory. It fits well and doesn't break any further conjectures{]}\newline ``Okay - I'll accept that 2 is a number. Then my conjecture is `All even numbers except 2 are the sum of two primes'.''
\end{itemize}
\end{ep}

Let us consider whether either of Episodes \ref{ex:platypus} and
\ref{ex:goldbach} meet our criteria.
The system's \emph{perception} again relies on its generative methods,
drawing where relevant on external systems.  Agents develop concepts,
conjectures, theorems, and examples that are given preliminary
assessments: the most interesting findings are shared during the
``discussion phase''.  This is reasonable evidence of
\emph{attention}.  By comparison with {\sf HR} in Episodes
\ref{ex:central} and \ref{ex:monoid},
context- and data-specific \emph{focus shifts} are integral to {\sf HRL}'s agent-based model.  This is because each agent is working with its own theory,
and can independently decide what to do with the evidence shared by the other agents.  New contexts are frequently in play due to the different agents working in slightly different spaces.

Thus,  in Episode \ref{ex:platypus}, \emph{A}'s statement ``There does not exist an animal which produces milk and lays eggs'' is initially recontextualised by \emph{B} and given a negative evaluation (since it is refuted by the existence of the platypus). Subsequently, however, \emph{A} considers the same statement in a new context in which the platypus has been deleted.  The statement is then given a positive evaluation.  In the course of this exchange, a \emph{focus shift} has taken place (satisfying both the precondition and condition \emph{(i)} of Definition \ref{def:focus-shift-ability}). The initial conjecture becomes true, because
the counterexample has been excluded.  HRL's \emph{explanation} for its answer that the existence of such an animal violates many conjectures. No further problem has been solved, however, because the original observation is identical to the final conclusion, hence there is no \emph{bridge} step.\footnote{For comparison, counterfactually de Mestral might have decided that cockleburs were inherently interesting, but never gone on to create Velcro\textsuperscript{\texttrademark}.} Externally, the \emph{evaluation} of the result is negative, as the system has ``undiscovered'' an actually-existing animal. (However, it it worth noting that when run under different parameters, HRL will ``discover'' the platypus, which receives a positive evaluation as a particularly interesting animal.) 

In Episode \ref{ex:goldbach}, the \emph{focus shift} step is more involved.
Suppose that $\mathcal{E}$ is the initial conjecture
``All even numbers are the sum of two primes.''
When \emph{B} finds a counterexample to $\mathcal{E}$, it is given a negative evaluation.
Agent \emph{B} subsequently supplies \emph{A} with a new context, now enriched with the
number $2$.  At this point, \emph{A} 
could in principle simply discard the initial conjecture, but it does not.
Instead, the conjecture is given an intermediate positive evaluation:
it is clearly incorrect, but it is still interesting.
Specifically, \emph{A} is able to employ Lakatosian ``piecemeal exclusion'' to
remove $2$ from the set of numbers covered by the conjecture, producing
$\mathcal{E}^\prime$, ``All even numbers except $2$ are the sum of two
primes.''

Here, Agent \emph{A} has combined
\emph{B}'s counterexample with the original conjecture, thereby forming
a \emph{bridge} to an interesting problem, Goldbach's conjecture.
The new conjecture is given a positive \emph{evaluation} by {\sf HRL}
for the same reason the conjecture is historically interesting: it is succinctly
stated, but continues to evade proof.  However, since the conjecture
is already well known (and remains unproved), the simple
fact of its reformulation by {\sf HRL} has no chance of 
receiving the kind of recognition given to original mathematical
discoveries---of the sort that have in fact been made with {\sf HR}
\citep{colton2007computational}.

\begin{table}
\begin{center}
  \resizebox{1\textwidth}{!}{%
  \begin{tabular}{l|L{.16\textwidth}|L{.16\textwidth}|L{.16\textwidth}|L{.16\textwidth}|L{.16\textwidth}|L{.16\textwidth}|} 
    \multicolumn{1}{l}{}&\multicolumn{1}{l}{\textbf{Object}} & \multicolumn{1}{l}{\textbf{Context 1}} & \multicolumn{1}{l}{\textbf{Eval.~1}} & \multicolumn{1}{l}{\textbf{Context 2}} & \multicolumn{1}{l}{\textbf{Eval.~2}} & \multicolumn{1}{l}{\textbf{Focus Shift}} \\[.3em]
    \cline{2-7}
    Ep.~\ref{ex:central}&Concept of central elements of a group&Background& $>\theta$&---&---&NO,\newline Precondition not met; {\em (i)-(iii)} not met \\
    \cline{2-7}
    Ep.~\ref{ex:monoid}&Boring conjecture in monoid theory &Background& $>\theta$ & Background + Counter-example & $<\theta$& NO,\newline Precondition not met (although {\em (ii)} is satisfied)\\
    \cline{2-7}
    Ep.~\ref{ex:platypus}&There does not exist an animal which produces milk and lays eggs. &\emph{B}'s Background (including platypus) & $<\theta$ & \emph{A}+\emph{B}'s background, with platypus deleted& $>\theta$ &YES,\newline Precondition and condition {\em (ii)} are met\\
    \cline{2-7}
    Ep.~\ref{ex:goldbach}&All even numbers are the sum of two primes.&\emph{B}'s background (including $2$) & $<\theta$ & \emph{A}'s background + 2 + ability to perform Lakatosian piecemeal exclusion & $>\theta$&YES,\newline Precondition and condition {\em (ii)} are met\\
    \cline{2-7}
  \end{tabular}}
\end{center}
\caption{\label{hr-episodes-table}Presence or absence of conditions for a focus shift in {\sf HR}/{\sf HRL} in Episodes 1 through 4. We use $\theta$ to represent an arbitrary threshold, with different values in each example (see Definition \ref{def:focus-shift-ability}).}
\end{table}

\section{Discussion} \label{sec:discussion}





The examples in the previous sections show that serendipitous behaviour can be exhibited in a meaningful sense by computer systems. 
The demonstration of this
claim has made use of a novel theoretical synthesis, which, nevertheless,
is compatible with other established perspectives on serendipity.  We are not
the first to argue that the potential for serendipity can be
increased---or, indeed, decreased---because of technological design
choices (e.g., \cite{danzico2010design}, \cite{newman2002designing}, \cite{melo2018}).
However, this seems to be the most comprehensive effort to date to relate
theories of serendipity to work in artificial intelligence.

The effort incorporates an ``ecological'' \citep{kenyon2013ecological}
perspective on artificial intelligence, in which the system develops
in relationship to its operating environment.  This bears on the
concept of ``self-improving'' \citep{Majot2017} AI systems.  The model
of serendipity potential details one way in which such improvements
can be structured.
Below, we discuss additional related work (Section \ref{sec:related}),
including existing research that incorporates or references our model
(Section \ref{sec:incorporating}), along with potential applications
in computational creativity research (Section
\ref{sec:cc-applications}).  First, we summarise the key implications
of the work presented above.

\subsection{Implications} \label{sec:implications}

Looking into the foundations of the focus shift, we must reject
theories of serendipity that rely entirely on blind selection
mechanisms, just as we must reject theories based on perfect control.
The word `blind' is understood to mean the complete absence of reliable
advance knowledge of benefits, rather than a specific perceptual
deficit.  One prototypical example is a radar system which scans in
360\textsuperscript{$\circ$} for ships or aeroplanes
\cite[p.~383]{campbell1960blind}; another is classical Darwinian
evolution.  Simonton cited {\sf BACON}  as
an example of a `blind' but nonetheless ``systematic'' discovery
system, based on ``heuristic methods''
\cite[p.~169]{simonton2010creative}.  Like {\sf HR}, {\sf BACON}'s
heuristics are implemented using production rules
\cite[p.~69]{langley1987scientific}.
Importantly, {\sf BACON}'s production rules:
\begin{quote}
``\emph{also incorporate information about the current goals of the
    system, so that a compromise between data-driven, bottom-up
    behavior and goal-driven, top-down behavior can be achieved.}'' \cite[p.~70]{langley1987scientific}
\end{quote}
Accordingly, Simonton's
analysis can be contrasted with Austin's \emph{`barking up the right
  tree' phenomenon}:
\begin{quote}
``\emph{if you happen to be the kind of person who hunts
afield, it may be, in fact, your dog who leads you up to the correct
tree, and to a desirable conclusion}'' \cite[p.~50]{austin1978chase}.
\end{quote}
Recall from Section \ref{sec:etymology} that {\sf BACON}'s namesake
was a pioneer of serendipitous thinking \emph{avant la lettre}.  The
examples in Section \ref{sec:system-analysis} show that a context
shift alone is not sufficient to bring about a focus shift.  The focus
shifts we examined included both a changing context and an increasing
evaluation score.  It would seem that richer understandings of a
context and its likelihood to yield epistemic value will aid
serendipity, so that the ability to focus shift is anything but
`blind'.  Evolutionary models that incorporate learning, per
\citet{baldwin-effect} would be the relevant ones here (see \cite{fontanari2017revival} for a
contemporary survey).

\citet{grace2015specific} contend that a
``generative act is serendipitous if the search process possessed no
specific intent to create that artefact or anything like it'' (p.~264), which is again similar to Campbell's theory of creativity as a process of blind variation and selective retention.
They understand `intent' to arise within ``the
iterative process of defining the creative task and solving it in
parallel,'' and they connect this notion with curiosity: a ``drive to
explore what the system has observed but not understood'' (\emph{ibid.}, p.~261).
Intentions, so construed, could quite readily surround an unindented event and influence
its interpretation, and even influence its likelihood of occurring in the first place.

For example, \citet{guise2010redefining} unpicks the myth surrounding the
invention of vulcanized rubber, remarking that ``discovery and
invention are rarely simple events'' (pp.~359--360).  Goodyear worked
at a time when many people were seeking to make profit from
manufacturing rubber goods.  As it happens his initial patent did not ``originally
claim curing rubber solely with heat'' (\emph{ibid}., p.~379).
The patent was reissued and changed in subsequent iterations.
Reframing his discovery as a eureka moment with broad conceptual
coverage helped give Goodyear and his inheritors increasing control via the reissued patents.
A clear implication of this story is that the way we model and manage intention, accident, and their combinations has real-world consequences.


If the phenomenon of `blindness' came in degrees, we might observe the
propagation of prediction errors in a system that works to reduce
surprise over the long term, as in predictive processing and active
inference accounts of cognition. For a survey enlarging on our brief
framing in Section \ref{sec:modelTerms}, see \cite{newen2018oxford}.
We note that creative drives have also been discussed within this framework
\citep{Clark2017}.  While we have not wedded our modelling approach to
theories of predictive processing and corresponding Bayesian
architectures, it is worth remarking that ``surprise'' is crucial in
those models.  A response to an error in prediction can either
motivate action---which ameliorates the error by bringing the world
into alignment with our predictions---or else motivate adaptation of
the predictive models themselves.

Systems with these abilities could potentially find themselves at odds
with predefined rules mandated by AI ethicists.
\citet{caliskan2017semantics} recommend ``the explicit
characterization of acceptable behavior'' and the ``explicit
instruction of rules of appropriate conduct.''  While it is good and
perhaps necessary to be explicit when computers are involved, it seems
unrealitic to expect any one set of rules, fixed in advance, to apply
cleanly and universally in all circumstances.  Our world involves questions whose
``conditions are very numerous and inter-complicated''
\cite[p.~710]{lovelace}.  It is replete with feedback loops.

The work developed by
\citet{loughran2018serendipity} and \citet{mccallum2018} on
serendipity in music and video production, respecively, suggests the
usefulness of ecological approaches in the creative sphere.
Rather than managing uncertainty by fixing rules
once and for all, it may be possible to constrain AI systems using the
same kind of adaptable institutions that we use to manage human
societies (cf.~\citet{corneli2016institutional}).
With one foot in the world of accidental circumstance, perhaps
serendipitous events can never be fully explained: however, our model, and
refinements and implementations thereof, will aid in its rigourous study.


\subsection{Related work} \label{sec:related}

We have focused on ``serendipity in the system,'' but
\citet{Edmonds1994} arrived at a similar perspective to ours by thinking about tools
that could support the serendipitous creativity of their users.  He
argued that studying support tools is a useful way to investigate a
broader question: how do machines interact with their operating
environment?  He draws the conclusion that ``we are bound to consider
open system models of the creative process rather than the closed ones
implied by the Turing Machine'' (p.~341).  Indeed, he points to
statements from Turing himself that
indicate the limits of the Turing Machine model, considering machines
that allow ``interference from outside,'' and in which ``such
interference is the rule rather than the exception'' \citep{turing1948intelligentreport}.  Elsewhere
Turing would use the convenient shorthand, \emph{learning machines}
\citep{turing1950mind}.  According to Turing's analysis, applications
such as language learning and human-level mathematics are likely to
require rich contact with the outside world.  Concerning the process
of learning mathematics, and with reference to Kantian foundations,
\citet[p.~2015]{sloman2008well} again highlights ``requirements
\ldots\ arising from interactions with a complex environment.'' 

\citet{swanson2016predictive}
indicates that the predictive processing framework, an
inspiration for our model, ``should not be regarded as a new paradigm, but
is more appropriately understood as the latest incarnation of an
approach to perception and cognition initiated by Kant and refined by
Helmholtz.''
Kant had contended that ``reason has insight only 
into what it itself produces according to its own design,''
and disparaged the notion of
learning from accidental observations absent ``a previously thought out
plan'' \cite[p.~20]{kant1929critique}.  One also wonders, just what can be learned from a previously
thought out plan in the absence of accidents?
Van Andel's insistence that pure serendipity
cannot be manifested by a computer program seems to address this question.
And yet, the hard line that he takes on the matter
might be tempered if the program in
question was allowed to implement a learning machine in the sense indicated by Turing.

In fact, Kant was also led to consider something akin to unsupervised learning,
which he called \emph{reflective
judgement}.  This process subsumes objects ``under a law
that is yet to be given \ldots\ under a law which is in fact
only a principle of reflection on objects for which we have no objective
law at all''
\cite[p.~265]{kant1987critique}.
This is compatible with the considerations above regarding previously thought out plans.
Reflection is
a ``subjective principle governing the purposive use of our cognitive powers'' \cite[p.~266]{kant1987critique}.  As an example along these lines, Eco suggested that,
had Kant had the opportunity to observe the platypus, he
would have concluded that it 
is ``a masterpiece of design, a fantastic example of environmental
adaptation, which permitted the mammal to survive and flourish in rivers''
\cite[p.~93]{eco2000kant}.
There is quite a difference between this creative
line of abductive reasoning and {\sf HRL}'s reductive approach, traced in
Section \ref{sec:pursuit}.  When platypus specimens were first exhibited in scientific circles, the creature was thought to be a hoax: {\sf HRL} partially reconstructs this reaction.  However, as we saw in that section, given
a somewhat richer background theory, {\sf HRL} was also capable of exercising
something akin to reflective judgement, and could thereby reinvent
a famous number-theoretic conjecture.


In Section \ref{sec:literature-summary}, we suggested that serendipity
is \emph{a form of creativity that happens in context, on the fly,
  with the active participation of a creative agent, but not entirely
  within that agent's control.}
We also remarked there that \citet{copeland2017serendipity} has argued
for a contextual perspective on serendipity that ``goes beyond the
cognitive.''  While our approach has centred on cognitively-plausible
computational modelling, we have  had in mind what
Edmonds referred to as ``open system models.''    The
perspective we developed in Section
\ref{sec:modelTerms}  is compatible with what
\citet{tonnessen2015uexkullian} calls ``Uexk\"ullian phenomenology.''
T{\o}nnessen's conception of a world rich in interdependence across
various layers of mental processing is also compatible with Copeland's
assertion that serendipity is found in networks and communities, and
in mundane social encounters.

 
While Copeland suggests that ``serendipity is a category that can
only be applied retrospectively to a discovery process'' 
\cite[p.~7]{copeland2017serendipity}, she also
mentions several skills and cultural traits that can be cultivated to
encourage serendipity, such as the early sharing of research results.
Although we have presented the steps of our model building on one
another in sequence, feedback loops are allowed, and 
experimentation with different architectures will be important.
Certain core features are needed.  In addition to the central
role played by the focus shift, the major phases of discovery and invention depicted in
Figure \ref{fig:model} amount to model-building and model use.
\citet[p.~720]{kockelman2011biosemiosis} contends that just as ``one
cannot offer an account of significance without an account of
selection'' also ``one cannot offer an account of selection without an
account of significance.''  In order to have serendipity potential,
systems need to model the anticipation and appreciation of
valuable outcomes in an uncertain world.

\citet{bjorneborn2017three} expands upon the theme of encouraging
serendipity in considerable detail.  He puts forward three major
``personal factors in serendipitous encounters'': \emph{curiosity}, \emph{mobility},
and \emph{sensitivity}.  These correspond to three parallel environmental
factors or affordances, which he terms ``diversifiability,
traversability, and sensorability.''  Both sides of this balance are
then described in terms of sub-factors, ten on each side.  However, while
Bj\"orneborn notices an interesting parallel between agent and
environment, he does not comment explicitly on a parallel with the
classic theory of mind in three parts, namely the
``\emph{conative},'' ``\emph{cognitive},'' and ``\emph{affective}''
\citep{hilgard1980trilogy}.
Links with the three personal factors mentioned can be readily
traced.
\citet[p.~347]{boden1998creativity} notes that creativity similarly
``involves not only a cognitive dimension (the generation of new
ideas) but also motivation and emotion.''
Two of Bj\"orneborn's sub-factors, sensitivity-attention and
curiosity-interest, show up as facets in our model.  However, the
three dimensions may be active more widely, which is why they were not
included in Table \ref{tab:theory-summary}.

Previous work  described an
information-processing model of \emph{insight} \citep{demystification},
after the outline provided by \citet{wallas1926art}.
Such ideas point to applications of
computational technology that ``facilitate the discovery of
previously unknown cross specialty information of scientific
interest,'' as discussed by
\citet[p.~183]{swanson1997interactive}, i.e.,
``literature-based discovery'' \citep{smalheiser2017rediscovering}.
In the approach of Swanson and Smalheiser, conditions of
\emph{complementarity} and \emph{noninteraction} between
two bodies of literature suggest the presence
of ``unnoticed useful information,'' which may be hinted at through
``indirect linkages'' \cite[pp.~184, 185]{swanson1997interactive}.
One class of explicit indirect links are \emph{bridging terms},
as mentioned in Section \ref{sec:modelTerms}.
Surfacing these connections drives at insight,
if that is understood to mean ``an improved representation
of an important previously unsolved problem, which now likely contains
the essence of a correct solution'' \cite[p.~118]{demystification}.

The broader parallels between Wallas's model
of creativity and contemporary receptions of the concept of serendipity
(Section \ref{sec:distill}) suggest that the latter concept goes beyond
insight.
Cases of true serendipity integrally involve what Swanson and Smalheiser refer to as ``problem generating''
\cite[p~.186]{swanson1997interactive}.  But in
serendipity, this happens relatively late in the process, rather
than at the outset as it did in Swanson and Smalheiser's work.  \citet[p.~153]{kulkarni1988processes} suggest a
related heuristic: ``If the outcome of an experiment violates
expectations for it, then make the study of this puzzling phenomenon a
task and add it to the agenda.''
By remaining \emph{open} \citep{jurvsivc2012cross} to the
identification and pursuit of new challenges,
potentially-serendipitous processes are able to pose and solve novel, useful,
problems.
All of this comes with significant demands for any implementation: our
examples have shown that these can be met, though we have also seen that
such an implementation may not convey immediate practical advantages.


To emphasise just what it is that serendipity in the system
could bring to the table,
consider the example of {\sf Max}, a system
designed to provide serendipity as a service
\citep{Figueiredo2001,campos2001searching}.
{\sf Max} modelled
users' interests as word vectors,
extracted from emails; these were
converted to conceptual structures using WordNet;
{\sf Max} then suggested new web pages for the
user to read.  
{\sf Max} was capable of delivering, albeit with
low probability, recommendations deemed to be
of considerable value.
Examples of both pseudoserendipitous and serendipitous
varieties were adduced \cite[p.~59]{Figueiredo2001}.
However, {\sf Max} was not open to discoveries in the sense described
above, and as such could not carry out
new use-inspired research to improve its performance.  For example,
{\sf Max} applied \emph{term frequency-inverse document frequency}
(tf-idf) to rank the concepts in each user-supplied document
\cite[p.~160]{campos2001searching}---but there is no chance that the
system, as architected, would decide to try reducing the dimensionality
of the associated vector space, and then use declustering (like {\sf
  Auralist} of \cite{Zhang2011}) to see if this improved recommendation
quality.  The conditions that led to the historically-significant
extension of tf-idf into \emph{latent
  semantic analysis} (LSA) are simply not modeled in {\sf Max}---even
though the program was built with a somewhat-similar problem in mind:
\begin{itemize}
\item \cite{landauer2003pasteur}, who pioneered LSA: ``the words that people wanted to
  use, to give orders to computers, or to look things up, rarely
  matched the words the computer understood.''
\item \cite{campos2001searching}, creators of {\sf Max}: ``Information retrieval usually
  assumes that the users know what they are searching for [but information can also be acquired] in an accidental, incidental, or
  serendipitous manner.''
\end{itemize}
Recent advances in reasoning about programmatic data structures (e.g., \cite{patterson2017,patterson2018})
may help accelerate the development of robust tooling that exhibits serendipity in the system.
We are aware of varied recent systems that make other interesting
innovations: some of these are mentioned below.

\subsection{Work that incorporates or references the model, and potential for further development} \label{sec:incorporating}

We can reflect in practical terms on Copeland's advice concerning the
sharing of early research results.  During the development of our model,
previous iterations of the paper have been made available via
arxiv.org \citep{arxiv} and discussed in two AISB symposia.
To date, 22 publications have
cited the working version of the paper on
arxiv.org,
which has given us an impression of how others think about the model.\footnote{\url{https://scholar.google.co.uk/scholar?oi=bibs&hl=en&cites=8190354202005420104&as_sdt=5}. Citation
  count accurate as of 8 March 2020.}

In their recent paper exploring serendipity in computer-generated
fiction, \citet{mccallum2018} reflect on how the detail in our model
``more clearly articulate[s] what must occur for the chance encounter
to be productive,'' which can help designers of AI systems take
advantage of the ``productive and perilous moment \ldots\ in which an
unexpected event or pattern occurs [that might otherwise go unnoticed
  or unrecognised]'' (p.~7).  
\citet{wopereis2017} remark that
modelling serendipity in computational systems is a topic that is
growing in interest: ``Seeking serendipity may sound as a paradox,
just like controlling it, [however there is] increasing evidence that
we can influence and stimulate it.'' \citet{surroca2015} noted that
our work was the only instance of ``the formalization and the
measurement of this phenomenon'' that they had knowledge of (p.~404).

Here we should stress that quantitative measurement of serendipity
potential, which we had attempted to deal with in an earlier draft of
this paper, gives rise to complications that have since caused us to
beat a retreat.  A full picture of serendipitous creativity must take
into account both the discoverer and the environment, and in the
valuation step, the discovery itself, if not also way it is
communicated (cf.~\cite{jordanous2016four}).  Measuring the
serendipity potential of a given system is not realistically possible
without knowing a great deal about the landscape in which that system
operates.  This does not detract from the possibility of
operationalising the concept of serendipity potential within specific
applications, as our analysis above shows, and as we detail in further
examples below.  It might be possible to formalise the concept of
serendipity potential in a Solomonoff-style probabilistic treatment
\citep{solomonoff1986application},
or as a suitably formulated Bayesian reinforcement learning problem \citep{vlassis2012bayesian},
or in some other framework, but this must be left for future
work.  In addition, while we have been inspired by predictive processing
and active inference, the project of formally redescribing the model
in terms of the situated, recursive, neural architectures frequently referred to
in that line of work is similarly deferred.

The existing model's qualitative aspects have informed discussions of
the serendipity potential of recommender systems
\citep{kotkov2016survey,patel2018} and the reporting of serendipitous
events \citep{allen2018}.  The framework was also referenced briefly in
connection with research into serendipity in revenue models
\citep{bechmann2016}, preference-guided content discovery on the Web
\citep{surroca2015}, computational models of curiosity
\citep{grace2017}, literary creativity \citep{gervas2016integrating} and
musical improvisation \citep{wopereis2017}.  All of these would be
interesting topics to develop further, and such investigations would
be likely to give rise to additional domain-specific heuristics. 

\subsection{Applications of computational serendipity within computational creativity} \label{sec:cc-applications}

Serendipity has been of considerable interest in computational
creativity research, where it has been discussed alongside other topics
like ``intention, recognition, and generation''
\citep{Pease_Jordanous_2018} that bear on the nexus of creativity and
discovery.

By now our comments adapting the notion of blind search are well
establish, so when \citet{veale2011reuse} remarks that ``serendipitous
discovery is unlikely to arise in purposeful explorations'' the usual
caveats are needed.  As per McKay's reading of Bergson, the creation of a large database of
photographs (\textsection\ref{sec:etymology})---or, in Veale's work, phrases---is not sufficient to bring
about serendipity.  However, breadth of experience is a
necessary aspect of the prepared mind, and a constituent of many forms of creativity.  Thus, for example, from its etymology the concept of `serendipity' is
itself almost a linguistic \emph{objet trouv\'ee} in the sense discussed by
Veale (\emph{op. cit.}, and more recently, \citet{veale2016grounded}).
Silver contends that
\begin{quote}
``\emph{it took a belletrist and sharp-witted dilettante to read Bacon
  as a champion of accident---despite the manifest commitment of
  Bacon's work to the establishment of method.}''
  \cite[p.~256]{Silver_2015}
\end{quote}
It is \emph{the ability to focus shift} that allows complex appropriations
and reinterpretations to become meaningful in a new context.

Modelling large corpus collections is an ongoing strand of
work within computational creativity research (e.g.,
\cite{McGregorEtAl15JAGI}).  Further development of the abilities
implied by our model must go beyond building models
of meaning, so long as those remain disconnected from practice.  From a
practical standpoint, it is important to emphasise that what we have been referring to as
serendipity in the system could be developed in symbiosis with
user-facing services. The role of the user has been discussed in connection with other machine
learning technologies \citep{amershi2014power}.  It may be natural to combine serendipity in the system with serendipity as a service.
Promising application areas range from education
\citep{mohseni2019Pique} to healthcare \citep{niu2018surprise} and
beyond.  From the point of view of our model, current serendipity
support tools miss the opportunity to work in a `virtuous serendipity circle.'  In
future tools, the system could simulaneously support the user's
experience of serendipity, and adapt to underlying changes in the domain or in user
behaviour to support the system's ongoing serendipitous development.  These remarks
are not merely speculative: though much remains to be done, there has been recent attention to
developing \emph{adaptive recommender systems} \citep{shengbo-guo-thesis,niu2018adaptive}
which would provide a natural point from which to build towards serendipity in the system.

Aesthetic domains also offer a range of application areas for models incorporating serendipity potential.
\citet{jordanous10} reported on a system using genetic algorithms for
computational jazz improvisation, which was later given the name {\sf
  GAmprovising} \citep{jordanous:12}.
\begin{quote}
\emph{``Over several runs, it was able to produce jazz improvisations which
slowly evolved from what was essentially random noise, to become more
pleasing and sound more like jazz to the human evaluator's ears''}
\citep{jordanous10}.
\end{quote}
\citet{kaliakatsos2016argument} used blending in a music context, but
as with {\sf GAmprovising}, their system required a human in the loop
for evaluation purposes.  More recently
\citet{loughran2018serendipity} drew on the concept of ``cybernetic
serendipity'' in their design of a music system driven by a
{``}`circular-causal' loop,{''} which employs a population of evolving
critics to build an emergent fitness function, which in turn guides
the evolution of melodies.  Here, the human programmer plays a more
abstract role.  As in other dimensions of computational creativity
\citep{stakeholder-groups-bookchapter}, we might see
progressively more responsibility for developing serendipity potential
handed over to the machine, as part of a trend towards increased autonomy.

With regard to Harold Cohen's painting program, \citet[p.~340]{Edmonds1994} remarks
``Perhaps the prime restriction on {\sf AARON}'s creativity is that it
cannot see.'' 
Although more recent computer painting programs have
overcome this limitation (e.g., in \citet{colton2015painting}), this
does not immediately translate into richly meaningful behaviour.
\citet{karimi2019sketching} describe a system that can perform \emph{conceptual shifts} that
involve ``viewing what has been drawn through a new conceptual lens.''  This is clearly a
promising direction for further work.

Pointing to a way to think about such conceptualisation,
\citet{gucklesberger2017addressing} characterise creativity
with a series of ``why questions'' that creative systems
would need to be able to address in order to explain their
behaviour convincingly.  At a higher level we can ask who is responsible for asking the driving questions.
For example, \citet{bou2015role} show that concept
blending can be applied to analyse and retrospectively reconstruct
mathematical examples, but that much more work would be needed to build a mathematical
system that convincingly asks questions which drive the selection of the
items to blend.

The ability to generate a cogent and socially meaningful explanation or rationale
will become especially important when the system could drastically change its behaviour
based on what it observes in the environment.  Gerv\'as describes the
classic system \textsf{Author}:
\begin{quote}
``\emph{Dehn postulates two different metagoals: achieving the current
  narrative goal and finding better narrative goals to pursue. It is
  this second metagoal that guarantees the directed-serendipitous
  duality, allowing for changes in direction when unforeseen
  opportunities arise.}'' \cite[p.~54]{Gervas_2009}
\end{quote}
Of course, \textsf{Author} and all other computer systems will have
limitations.  We have developed an outline showing how these can limits be pushed further.  Here the outlook is positive.  Corresponding risks associated with computational
systems that can change their goals have been frequently discussed in works of
science fiction.  We believe that, by and large, that is where such discussions
belong.  This is not to say that such fictional works have no purpose.
Our hope is that the model we have presented will help future scholars and the machines they employ exploit what might be termed Walpole's method,  ``leaving the powers of fancy at liberty to expatiate through the boundless realms of invention'' \citep[p.~xiv]{walpole1766castle}.

\section{Conclusions} \label{sec:conclusion}

Rich functional models of operating domains will be necessary for
  systems to recognise their own best and most interesting efforts, to
  identify new problems, and to exploit serendipitous outcomes when
  they occur.  Referring to some of the examples we examined, while
  {\sf DAYDREAMER} met the basic requirements of our framework, it
  does not have a robust way to discriminate between more and less
  interesting daydreams; nor can it adjust its view on the world to
  take in new perceptions based on its creative process.  Similarly, while {\sf HRL} met the basic requirements
  of our framework, to make discoveries of significant value it would
  need to be revised to draw on a wider range of scientific and
  mathematical knowledge.  {\sf Max}
  could potentially scaffold the user's experience of serendipity,
  but was not open to considerably shifting its own terms of
  engagement.  This critique suggests interesting
  directions for further work.

Current thinking about AI policy points out considerations related to
verification, validity, security and control that can reduce the
incidence of surprising behaviour in such systems
\citep{research-priorities}, but, so far, less attention has been given
to features that would allow autonomous systems to make beneficial use of
surprises they encounter.  This highlights an all-too-human bias, rather than
an objective limitation of machines.

The individual components of our model of serendipitous processing
have been supported with references to both classic and contemporary
systems.  Taken as a whole, the model addresses learning, adaptation,
and creativity in contexts with unpredictable features.
The model is effective at showing evidence for or against the
serendipity potential of existing systems.

The heuristics that we described can inform future implementation
and evaluation work.
For reference, an outline summarising the theoretical
foundations and heuristics from Section \ref{sec:our-model} is
collected in Table \ref{tab:collected-heuristics}. 
Serendipity potential can be encouraged in computational systems:
further research may give more evidence as to when it should be
encouraged.  \citet{pease2013discussion} suggested to ``proceed with
caution in this intriguing area.''  The current paper offers a
considered view of the issues at stake.

\begin{table}
\begin{tabular}{>{\raggedright}p{\textwidth}}
\textbf{System-environment relationships differ widely, and develop}\\
\hspace{.5cm}\textbf{differently.} \\
\textbf{Chance can play various roles in shaping perception.} \\
\textbf{The system has limited control.} \\
To create the possibility for varied patterns of inference to arise, support\\
\hspace{.5cm}rich interfaces.  \\
To reduce constraints, allow features to be defined inductively. \\
Organise and process perceptions differently depending on the tasks\\
\hspace{.5cm}undertaken.  \\
\textbf{Adaptive attention is related to surprise.} \\
\textbf{Learning, context, and meaning begin to arise together with}\\
\hspace{.5cm}\textbf{attention.} \\
\textbf{To some approximation, features of the environment will be}\\
\hspace{.5cm}\textbf{attended to.} \\
Attention can be understood as competition for scarce processing resources.  \\
Attention can be time-delineated.  \\
Competition may be less natural when we can take advantage of parallelism.  \\
\textbf{Assess the data's potential for strategic usefulness.} \\
\textbf{Interest is related to curiosity.} \\
\textbf{Context change is a possible basis for belief revision.} \\
Interest can be linked to novelty in order to inspire learning.  \\
Interest can be linked to aesthetics in order to capture varied notions of\\
\hspace{.5cm}fitness.  \\
Beauty is in the eye of the beholder.  \\
\textbf{A new model yields an improved ability to make a prediction.} \\
\textbf{There are different kinds of viable explanations.} \\
\textbf{The system creates an explanation of the event for itself.} \\
Experiments can have limited scope and still be useful.  \\
Given a sufficiently rich background, only a small amount of new data is\\
\hspace{.5cm}needed.  \\
Learning is less efficient, but more widely applicable, than knowing. \\
Communication between agents can transfer causal information.  \\
\textbf{It is sometimes necessary or desirable to go beyond explanation.} \\
\textbf{Two cases: pseudoserendipity versus true serendipity.} \\
\textbf{The bridge is transformational.} \\
\textbf{A good problem can be identified by working at a meta-level.} \\
Similarity, analogy, and metaphor can be used to retrieve known problems.  \\
Concept blending may, but does not necessarily, help identify new problems.  \\
Working across domains can give rise to intriguing ideas.  \\
Experiments can give surprising insights.  \\
\textbf{Affection is based on reflection.} \\
Model a sense of taste. \\
Allow the system to use the world. \\
Allow the system to shape its own goals. 
\end{tabular}
\caption{Summary of theoretical foundations (in bold) and heuristics for implementation from Section \ref{sec:our-model}\label{tab:collected-heuristics}}
\end{table}

\section*{Acknowledgements}

Patrick Doherty, Thomas Baruzzi, and several anonymous reviewers commented on earlier versions of the paper.
We are grateful to the Society for the study of Artificial Intelligence and Simulation of Behaviour (AISB) for supporting two workshops
where participants engaged with this material, and to Yasemin J.~Erden (St.~Mary's
University) for further on-the-ground support at the first workshop that went beyond the
call of duty. Thanks to workshop participants
Mark Nelson, Claudia Chirita, Diarmuid O'Donoghue, Jasia
  Reichardt, Pek van Andel, Colin Johnson, Elaine O'Hanrahan, Eilidh
  McKay, Abigail McBirnie, Stephann Makri, and Lorenzo Lane; and
to Katie McCallum, Majed Al-Jefri, Kate Monson, Alexsandar
  Zivanovic, the estate of Edward Ihnatowicz, R\'ois\'in Loughran,
  Michael O'Neill, Dave Murray-Rust, Benjamin Bach, Ian Helliwell, Paul Melo, Miguel Carvalhais, Deitmar
  K\"oring, Paul Pangaro, Liss C. Werner, and Elaine O'Hanrahan (again).

\bibliographystyle{spbasic}

\end{document}